%% file: main.tex
\documentclass[11pt]{article}

\usepackage[preprint]{acl}
\usepackage{times}
\usepackage{latexsym}
\usepackage[T1]{fontenc}
\usepackage[utf8]{inputenc}

\usepackage[table,xcdraw]{xcolor}
\usepackage{microtype}
\usepackage{graphicx}
\usepackage{adjustbox} 
\usepackage{booktabs}
\usepackage{tikz} 
\usepackage{algorithm}
\usepackage{algorithmic}
\renewcommand{\algorithmiccomment}[1]{\{{#1}\}}
\usepackage{amsmath}
\usepackage{longtable}
\usepackage{mdframed}
\usepackage{tcolorbox}
\usepackage{subcaption}
\usepackage{enumitem}
\usepackage{multirow}
\usepackage{float}
\usepackage{array}

\usetikzlibrary{shapes, arrows.meta, shadows, calc}
\tcbuselibrary{skins}

\definecolor{promptbg}{HTML}{F7F7F7}
\definecolor{promptframe}{HTML}{4A6FA5}
\definecolor{prompttitle}{HTML}{2C3E6B}
\definecolor{examplebg}{HTML}{FAFCF5}
\definecolor{exampleframe}{HTML}{5B8C5A}
\definecolor{exampletitle}{HTML}{3B5E3A}
\definecolor{phasered}{HTML}{B03A2E}
\definecolor{phaseblue}{HTML}{2471A3}

\newtcolorbox{promptbox}[1]{
  enhanced,
  colback=promptbg, colframe=promptframe,
  coltitle=white, fonttitle=\bfseries\small,
  title={#1}, boxrule=0.6pt, arc=2pt,
  left=6pt, right=6pt, top=4pt, bottom=4pt,
}
\newtcolorbox{examplebox}[1]{
  enhanced,
  colback=examplebg, colframe=exampleframe,
  coltitle=white, fonttitle=\bfseries\small,
  title={#1}, boxrule=0.6pt, arc=2pt,
  left=6pt, right=6pt, top=4pt, bottom=4pt,
}
\newtcolorbox{defbox}{
  enhanced,
  colback=promptframe!5, colframe=promptframe!70,
  boxrule=0.5pt, arc=2pt,
  left=6pt, right=6pt, top=4pt, bottom=4pt,
  before upper={\setlength{\parskip}{4pt}},
}
\newtcolorbox{remarkbox}{
  enhanced,
  colback=exampleframe!5, colframe=exampleframe!50,
  boxrule=0.5pt, arc=2pt,
  left=6pt, right=6pt, top=4pt, bottom=4pt,
  before upper={\setlength{\parskip}{4pt}},
}
\newtcolorbox{propbox}{
  enhanced,
  colback=phasered!4, colframe=phasered!50,
  boxrule=0.5pt, arc=2pt,
  left=6pt, right=6pt, top=4pt, bottom=4pt,
  before upper={\setlength{\parskip}{4pt}},
}
\newtcolorbox{iterbox}[1]{
  enhanced,
  colback=white, colframe=promptframe!40,
  boxrule=0.4pt, arc=2pt,
  left=4pt, right=4pt, top=3pt, bottom=3pt,
  fonttitle=\scriptsize\bfseries\color{promptframe},
  title={#1},
  coltitle=promptframe,
  colbacktitle=promptframe!8,
  attach boxed title to top left={yshift=-2mm, xshift=3mm},
  boxed title style={boxrule=0.3pt, arc=1pt, colframe=promptframe!40},
}

\newcommand{\achead}[1]{\par\vspace{3pt}\noindent\textbf{#1}\hspace{0.4em}}

\usepackage{amsmath}
\usepackage{amssymb}
\usepackage{mathtools}
\usepackage{amsthm}

\usepackage[capitalize,noabbrev]{cleveref}

\theoremstyle{plain}
\newtheorem{theorem}{Theorem}[section]
\newtheorem{proposition}[theorem]{Proposition}

\theoremstyle{definition}
\newtheorem{definition}[theorem]{Definition}

\theoremstyle{remark}

\usepackage[textsize=tiny]{todonotes}

\title{Self-reflecting Large Language Models: A Hegelian Dialectical Approach}

\author{
  \mdseries
  Sara Abdali\textsuperscript{1}, Michael Solodko\textsuperscript{1}, Can Goksen\textsuperscript{1}, Saeed Amizadeh\textsuperscript{1}, \\
  Julie E. Maybee\textsuperscript{2}, Kazuhito Koishida\textsuperscript{1}, Pashmina Cameron\textsuperscript{1} \\
  \textsuperscript{1}Applied Sciences Group (ASG), Microsoft, Redmond, WA, USA \\
  \textsuperscript{2}Department of Philosophy, Lehman College, City University of New York, Bronx, NY, USA \\
  \texttt{saraabdali@microsoft.com}
}

\begin{document}
\maketitle

\begin{abstract}
In this paper, we introduce a self-reflection framework for Large Language Models (LLMs) grounded in the \textit{Hegelian Dialectic}, a philosophical method in which an initial proposition is challenged by a generated opposition, and both are reconciled into a unified, more comprehensive idea. We formalize this process as an iterative operator over the space of consistent theories and apply it to two complementary tasks: \emph{(i)}~generating novel scientific ideas across domains such as mathematics, physics, economics, and philosophy, and \emph{(ii)}~improving reasoning by enabling LLMs to identify and correct their own errors through structured self-critique.
We study generation temperature through two configurations (a dynamic annealing schedule that shifts from creative exploration to refinement, and a constant temperature), to examine the effect of fixed versus dynamic temperature rather than advocate either. To evaluate ideas without domain experts, we introduce Multi-Agent Majority Voting (MAMV), in which multiple LLMs independently assess the validity and novelty of each synthesis. Our experiments show significant gains over baselines on mathematical (GSM-8k, GSM-hard), symbolic (GSM-Symbolic), and knowledge-intensive (MMLU Pro) reasoning, with promising qualitative results in open-ended scientific ideation.
\end{abstract}

\section{Introduction}
\label{Introduction}
LLMs have driven rapid progress in reasoning, scientific discovery, and creative problem solving~\cite{zhang2024comprehensive,wu2023impact,zhang2024scientific,zheng2023largelanguagemodelsscientific}, with In-Context Learning (ICL)~\cite{dong-etal-2024-survey} enabling zero- and few-shot reasoning from instructions or demonstrations without additional training. Yet ensuring factual accuracy and logical consistency in multi-step reasoning remains challenging~\cite{kddAIgeneratedtext,abdali2024llmsfooled}: LLMs neglect key conditions, misinterpret context, and hallucinate~\cite{Shayegani2023SurveyOV,milliere2023part1,abdali2024securinglargelanguagemodels}, which is especially costly in domains like mathematics and science where one logical error can invalidate an entire chain.

\begin{figure*}[t!]
\centering
\begin{tcolorbox}[enhanced, colback=white, colframe=promptframe, boxrule=0.8pt, arc=3pt, shadow={1.5pt}{-1.5pt}{0pt}{black!15}, left=2pt, right=2pt, top=2pt, bottom=2pt]
\includegraphics[width=\linewidth]{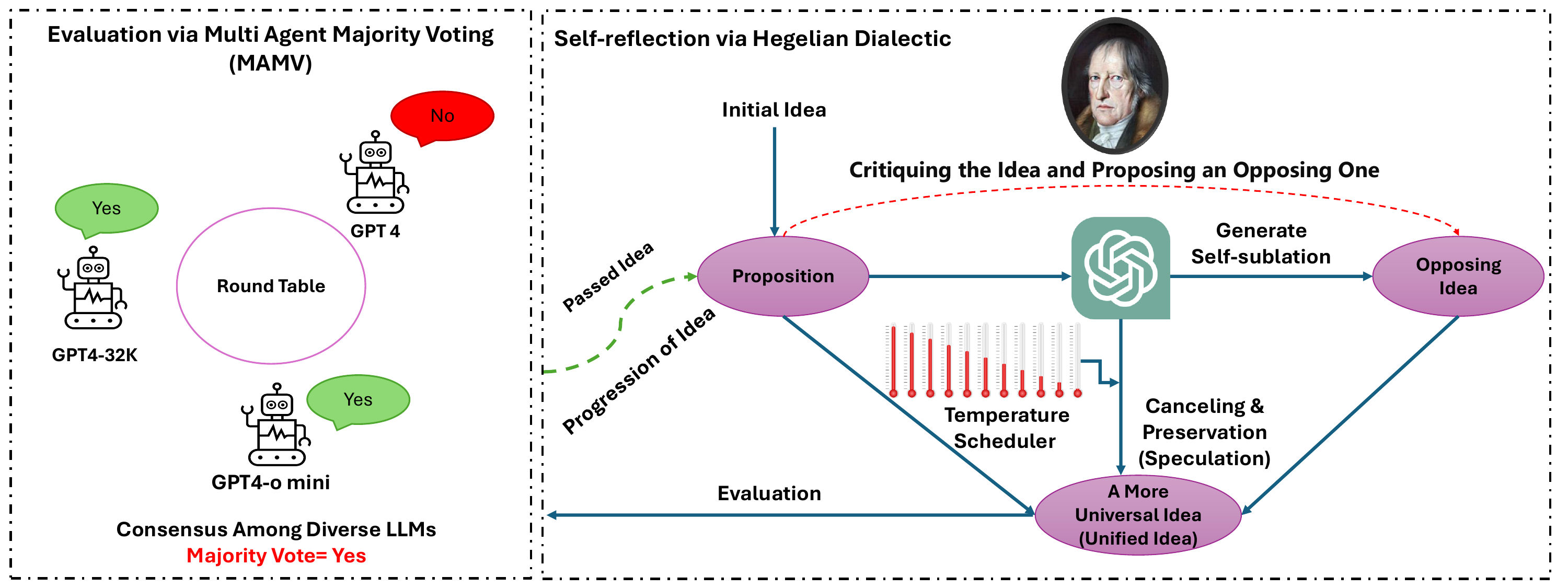}
\end{tcolorbox}
\vspace{-4pt}
\caption{\textbf{Overview of Hegelian Dialectical Self-Reflection.} A proposition is challenged via sublation and reconciled via speculation under two temperature configurations. MAMV evaluates novelty and validity of ideation.}
\label{fig:crownjewel}
\end{figure*}

\par Many strategies aim to mitigate this. \emph{Training-based} approaches include fine-tuning on curated data~\cite{lewkowycz2022solving,rajani-etal-2019-explain,zelikman2022star}, RLHF~\cite{Ziegler2019FineTuningLM,NIPS2017_d5e2c0ad}, and truthfulness-oriented data pruning~\cite{christiano2023deep}. \emph{Inference-time} methods include Chain-of-Thought (CoT)~\cite{wei2022chain}, Reversing CoT~\cite{Xue2023RCOTDA}, self-consistency~\cite{wang2023selfconsistency}, scratchpads~\cite{Cobbe2021TrainingVT,nye2022show}, and retrieval~\cite{Guu2023RALM}. More recently, \emph{multi-agent debate} (MAD) frameworks, Society of Minds~\cite{Du2023ImprovingFactuality}, Multi-Persona~\cite{liang2024encouragingdivergentthinkinglarge}, and ChatEval~\cite{chan2023chatevalbetterllmbasedevaluators}, show that structured argumentation between agents improves factual accuracy in QA~\cite{smit2024goingmadlookmultiagent}.

\par A complementary direction is \emph{iterative self-reflection}~\cite{shinn2023reflexion,madaan2023selfrefine}, where a model critiques and improves its own outputs. However, naive self-reflection suffers from \emph{degeneracy-of-thought}: once converged on a high-confidence answer, later iterations add no meaningful revision~\cite{liang2024encouragingdivergentthinkinglarge,wang2024unleashingemergentcognitivesynergy}, motivating a more structured form of self-critique.

\par We view self-reflection as a form of \emph{self-dialectic}, a discourse that advances by considering and resolving opposing views~\cite{CambridgeDialectic}, with meaning varying across traditions~\cite{sep-dialectical-school, sep-hegel-dialectics}. In particular, the \textit{Hegelian Dialectic}~\cite{hegel_phenomenology_of_spirit, Hegel1951-HEGHSO, sep-hegel-dialectics} has an initial proposition generate its own opposition by exposing its defects, then reconcile the two into a higher-order unified idea preserving the strengths of both. This structure (understanding, sublation, speculation) overcomes degeneracy-of-thought, as it \emph{requires} a substantive opposition before any synthesis.

\par Philosophy-inspired NLP has attracted growing interest, connecting computational methods with classical methodologies~\cite{milliere2023part1,milliere2024part2}. We propose a Hegelian dialectical approach to structured self-reflection that enables LLMs to generate novel ideas, identify errors, and correct them during problem solving. We formalize the dialectic as an iterative operator over consistent theories (Section~\ref{Hegelian Dialectical Method}), instantiate it with LLM-based sublation and speculation operators (Section~\ref{alg:LLM-self-reflection-llm-relevance-novelty}), and explore the effect of generation temperature via two configurations: a \emph{dynamic annealing} schedule formalizing the shift from creative exploration to focused refinement~\cite{Delahaye2019}, and a \emph{constant temperature} setting, treating both as configurations to study rather than proposing either as our method. To evaluate generated ideas without domain experts, we introduce \textit{Multi-Agent Majority Voting (MAMV)}~\cite{minsky1988society,Zhuge2023MindstormsIN,Amirizaniani(ToM)}, in which multiple LLMs independently vote on validity and novelty. Figure~\ref{fig:crownjewel} overviews our method. In summary:\vspace{-2pt}
\begin{itemize}[noitemsep, topsep=0pt, parsep=0pt]
    \item We introduce a self-reflection framework grounded in Hegel's dialectic, formalized as an iterative operator with provable consistency, preservation, and progress properties.
    \item We study the effect of fixed vs.\ dynamic temperature through extensive analysis.
    \item We propose MAMV, a scalable framework to assess validity/novelty without experts.
    \item We achieve significant gains across multiple models and benchmarks, plus promising scientific ideation across topics.
\end{itemize}
\vspace{-4pt}
\section{Related Work}
\label{Related Work}
\paragraph{Self-Reflection in LLMs}
In the context of LLMs, self-reflection involves evaluating and refining a model's outputs through iterative cycles of feedback and adjustment~\cite{chen2024selfcognition,li2024selfcontrast}. Li et al.~\cite{li2024selfcontrast} explore diverse perspectives, summarize discrepancies, and organize them into a checklist for refinement. Madaan et al.~\cite{madaan2023selfrefine} improve LLM outputs through iterative feedback, while Reflexion~\cite{shinn2023reflexion} enables agents to learn from trial-and-error. Self-reflection has also been applied to Retrieval Augmented Generation~\cite{zhang2024selfrag}, and translation~\cite{wang2024taste}. However, naive self-reflection, where the model is simply instructed to reconsider its answer, suffers from \emph{degeneracy-of-thought}: after converging on a high-confidence response, the model fails to diverge in subsequent iterations~\cite{liang2024encouragingdivergentthinkinglarge,wang2024unleashingemergentcognitivesynergy}. Our work addresses this limitation by replacing unconstrained self-critique with \emph{structured dialectical opposition}, ensuring that each iteration introduces a substantive counterargument grounded in identifiable~defects.
\paragraph{Philosophy-Inspired NLP}
Researchers have recently examined NLP through a philosophical lens, connecting computational methods with classical debates about cognition, semantics, and intelligence~\cite{milliere2023part1,milliere2024part2}. Philosophical prompting techniques have emerged as a means to enhance LLM reasoning: the \emph{Socratic method}~\cite{chang2023prompting} uses structured questioning to stimulate critical thinking, while the Kantian-inspired \emph{UPAR} framework (Understand, Plan, Act, Reflect)~\cite{geng2023upar} structures prompting to emulate human cognitive processes. Our work extends this line of research by drawing on Hegel's dialectical method, a richer philosophical framework that, unlike Socratic questioning or Kantian reflection, provides a formal mechanism for \emph{opposition generation} and \emph{speculative unification}, naturally suited to iterative refinement in generative models. To our knowledge, ours is the first to formalize Hegel's dialectic as a computational operator and instantiate it within an LLM-based reasoning framework.
\section{Background}
\label{Hegelian Dialectical Method}
\paragraph{Hegelian Dialectical Method}
The term \emph{Dialectic} denotes a logical discourse that advances through the consideration and resolution of opposing views~\cite{CambridgeDialectic}, a concept central to Hegel's dialectic (1807)~\cite{hegel_phenomenology_of_spirit,hegel2010science}. Unlike Plato's \emph{elenchus}, where a contradiction is terminal and forces the premises to be rejected, Hegel treats contradictions as \emph{productive tensions} that drive ideas toward higher unity and synthesis~\cite{popper1940dialectic,sep-hegel-dialectics}. And unlike Fichte's rigid thesis--antithesis--synthesis triad, Hegel described an organic process of \emph{self-sublation} (\textit{Aufhebung}) in which each stage simultaneously negates, preserves, and elevates the preceding one~\cite{sep-hegel-dialectics}.

\paragraph{From Philosophy to Computation}
This method is naturally suited to computational instantiation: it is \emph{iterative} (each cycle's output seeds the next, mirroring auto-regressive generation), \emph{constructive} (generates new content via speculative unification rather than merely eliminating, as in elenchus), and \emph{self-correcting} (the sublation step explicitly surfaces defects, addressing the degeneracy-of-thought problem of naive self-reflection~\cite{liang2024encouragingdivergentthinkinglarge}). These motivate
our formalization.

\paragraph{Formal Characterization}
We ground the dialectical process in first-order logic (FOL), following the formalization of Inoue~\cite{inoue2014dialectical}. Let $\mathcal{L}$ denote a first-order language and let $\mathrm{Th}(\mathcal{L})$ be the set of consistent theories expressible in~$\mathcal{L}$. A \emph{dialectical trajectory} is a sequence $\langle T_0, (A_0, S_0), T_1, (A_1, S_1), \ldots, T_N \rangle$ where each $T_i$ is a proposition, $A_i$ is its opposition, and $S_i = T_{i+1}$ is the speculative unification.

\begin{defbox}
\tiny\setlength{\parskip}{1pt}
\begin{definition}[Dialectical Opposition~\cite{inoue2014dialectical}]
\label{def:dialectical-opposition}
Let $L_1$ and $L_2$ be FOL languages, and let $T_1 \in \mathrm{Th}(L_1)$, $T_2 \in \mathrm{Th}(L_2)$ be consistent theories. For a sentence $w \in L_1 \cap L_2$:
\begin{enumerate}[label=(\roman*), nosep, leftmargin=*]
\item $w$ \emph{dialectically opposes} $T_2$ relative to $T_1$ \;iff\; $T_1 \vdash w$ \;and\; $T_2 \not\vdash w$.
\item $w$ \emph{dialectically contradicts} $T_2$ relative to $T_1$ \;iff\; $T_1 \vdash w$ \;and\; $T_2 \vdash \neg w$.
\end{enumerate}
We write $T_1 \underset{w}{\rightleftharpoons} T_2$ for opposition and $T_1 \underset{w}{\bowtie} T_2$ for contradiction.
\vspace{-3pt}
\end{definition}
\end{defbox}

\begin{defbox}
\tiny\setlength{\parskip}{1pt}
\begin{definition}[Dialectical Operators]
\label{def:dialectical-operators}
Let $\mathcal{L}$ be a first-order language and $\mathrm{Th}(\mathcal{L})$ the set of consistent theories in~$\mathcal{L}$. We define:
\begin{enumerate}[label=(\roman*), nosep, leftmargin=*]
\item \emph{Sublation}: $\mathcal{A}\!: \mathrm{Th}(\mathcal{L}) \!\to\! \mathrm{Th}(\mathcal{L})$ maps $T_i$ to $A_i\!=\!\mathcal{A}(T_i)$ s.t.\ $\exists w\!:\, T_i \vdash w \,\wedge\, A_i \vdash \neg w$.
\item \emph{Speculation}: $\mathcal{S}\!: \mathrm{Th}(\mathcal{L})^2 \!\to\! \mathrm{Th}(\mathcal{L})$ maps $(T_i, A_i)$ to a unified theory $T_{i+1}\!=\!\mathcal{S}(T_i, A_i)$.
\item \emph{Dialectical}: $\mathcal{D} = \mathcal{S} \circ (\mathrm{id} \times \mathcal{A})$, i.e.:
\end{enumerate}
\begin{equation}
\label{eq:dialectic-operator}
T_{i+1} = \mathcal{D}(T_i) = \mathcal{S}\bigl(T_i,\, \mathcal{A}(T_i)\bigr)
\end{equation}

\end{definition}
\end{defbox}

\begin{propbox}
\tiny
\begin{proposition}[Speculation Constraints]
\label{prop:speculation}
Let $S_{i} = \mathcal{S}(T_i, A_i)$. Then $\mathcal{S}$ satisfies $\forall\, T_i, A_i \!\in\! \mathrm{Th}(\mathcal{L})$:
\begin{enumerate}[label=(\alph*), nosep, leftmargin=*]
\item \textbf{Consistency:} $S_{i} \not\vdash \bot$\; {\tiny(non-trivial)}
\item \textbf{Preservation:} $S_{i} \supseteq T_i \cap A_i$\; {\tiny(common ground retained)}
\item \textbf{Progress:} $S_{i} \not\subseteq T_i \,\wedge\, S_{i} \not\subseteq A_i$\; {\tiny(advances beyond either)}
\end{enumerate}
\end{proposition}
\end{propbox}

\noindent The iterative application of~$\mathcal{D}$ generates a dialectical trajectory $\{T_0, T_1, T_2, \ldots\}$ with monotonically non-decreasing common ground between successive proposition--opposition pairs:
\begin{equation}
\label{eq:monotonicity}
|T_0 \cap A_0| \;\leq\; |T_1 \cap A_1| \;\leq\; \cdots
\end{equation}
This monotonicity reflects the accumulation of reconciled premises: as the dialectic progresses, a unified theory $T_{i+1}$ inherits an increasingly large set of validated claims from prior iterations, progressively narrowing the space of admissible oppositions while broadening the scope of the unification.

\begin{remarkbox}
\tiny
\textbf{Remark} (Non-determinism and Temperature).
\label{rem:nondeterminism}
The speculation operator $\mathcal{S}$ is \emph{not} uniquely determined. Multiple valid unifications may exist for a given pair $(T_i, A_i)$, forming a feasible set:
\[
\mathcal{F}_i = \bigl\{T \in \mathrm{Th}(\mathcal{L}) \;\big|\; T \supseteq T_i \cap A_i,\;\; T \not\vdash \bot\bigr\}
\]

In our LLM instantiation, we sample from $\mathcal{F}_i$ via the generation temperature $\tau$, which governs the entropy of the token-level distribution:
\begin{itemize}[nosep, leftmargin=*]
\item \textbf{Low $\tau$} (exploitation): concentrates probability mass on the most likely unification.
\item \textbf{High $\tau$} (exploration): encourages diverse speculative syntheses.
\end{itemize}
This connects the philosophical notion of \emph{speculative freedom} to the information-theoretic concept of \emph{distributional entropy}.
\end{remarkbox}

\paragraph{Hegel's Three Moments}
Hegel frames the process as three ``moments'' that map directly onto our operators~\cite{sep-hegel-dialectics,hegel_phenomenology_of_spirit}: \textbf{Understanding} ($\mathfrak{U}$), the stable current proposition $T_i$; the \textbf{Dialectical moment / Sublation} ($\mathfrak{A}$), in which the flaws of $T_i$ \emph{itself} drive an opposition $A_i=\mathcal{A}(T_i)$ (hence \emph{self}-sublation, which we enforce by conditioning $\mathcal{A}$ on $T_i$ alone); and the \textbf{Speculative moment} ($\mathfrak{S}$), a unified theory $T_{i+1}=\mathcal{S}(T_i,A_i)$ that integrates and transcends both, deriving its character from their specific opposition. The cycle is:
\begin{equation}
\label{eq:three-moments}
\underbrace{T_i}_{\mathfrak{U}} \;\xrightarrow{\,\mathcal{A}\,}\; \underbrace{A_i}_{\mathfrak{A}} \;\xrightarrow{\,\mathcal{S}\,}\; \underbrace{T_{i+1}}_{\mathfrak{S}}
\end{equation}
Crucially, $\mathfrak{S}$ sublates \emph{both} prior moments, negating them with a new concept yet still depending on them for its definition~\cite{sep-hegel-dialectics}.

\paragraph{Consistency and Paraconsistency}
A distinctive feature of the speculative moment is that oppositions need not be fully resolved, admitting ambiguity among competing perspectives~\cite{Hegel1951-HEGHSO,hegel_phenomenology_of_spirit}. Classically, $T_i \cup A_i$ may be inconsistent ($\vdash \bot$), which by the \emph{principle of explosion} would trivialize the theory ($\vdash \psi$ for \emph{any} $\psi$); we avoid this via the constraints of Proposition~\ref{prop:speculation}, resolving oppositions through qualification while accumulating their generating premises. This unresolved-opposition stance drew criticism from Popper~\cite{popper1940dialectic} and Russell~\cite{Russell1959-RUSMPD-6} for violating non-contradiction~\cite{sep-aristotle-noncontradiction}, but \emph{paraconsistent logics}~\cite{inoue2014dialectical,sep-hegel-dialectics} admit both $\varphi$ and $\neg\varphi$ without explosion ($\varphi,\neg\varphi \not\vdash \psi$), providing a natural foundation for Hegel's unifying step and framing our consistency constraint (Prop.~\ref{prop:speculation}a) as a design choice rather than a necessity.
\paragraph{Applications of the Dialectic}
The dialectic has been applied across politics~\cite{Boukhatem2022}, economics~\cite{HegelEconomicsBusiness}, physical sciences (e.g., geocentric vs.\ heliocentric astronomy, classical vs.\ relativistic mechanics)~\cite{inoue2014dialectical}, and ecological dynamics~\cite{ANGELER202539}, in each case synthesizing conflicting hypotheses into more comprehensive models.

\section{Proposed Hegelian Self-reflection}
Following the Hegelian Dialectic, an LLM $\mathcal{M}$ iteratively critiques its own \emph{Proposition} via a generated \emph{Opposition} and fuses both into a \emph{Unified response}.
\begin{figure}[!t]
\centering
\begin{tcolorbox}[enhanced, colback=white, colframe=promptframe, boxrule=0.8pt, arc=3pt, shadow={1.5pt}{-1.5pt}{0pt}{black!15}, left=6pt, right=6pt, top=8pt, bottom=6pt]
\centering
\scalebox{0.7}{%
\begin{tikzpicture}[xshift=8pt,
    node distance=1.5cm, auto,
    dialectnode/.style={
      draw, rectangle, rounded corners=5pt,
      minimum width=2.2cm, minimum height=0.85cm,
      font=\small\bfseries, text=white, inner sep=4pt,
      drop shadow={shadow xshift=1pt, shadow yshift=-1pt, opacity=0.3},
    },
    momentlabel/.style={
      font=\tiny\bfseries, text=white, fill=#1, rounded corners=2pt,
      inner sep=2pt, anchor=south,
    },
  ]
    \node [dialectnode, fill=promptframe] (being) {\shortstack{$T_i$: Proposition\\[1pt]\footnotesize(e.g., \emph{Being})}};
    \node [dialectnode, fill=phasered, right of=being, node distance=4.2cm] (nothing) {\shortstack{$A_i$: Opposition\\[1pt]\footnotesize(e.g., \emph{Nothing})}};
    \node [dialectnode, fill=exampleframe, below of=being, node distance=4.2cm, xshift=2.1cm] (becoming) {\shortstack{$T_{i+1}$: Unified Idea\\[1pt]\footnotesize(e.g., \emph{Becoming})}};

    \node [momentlabel=promptframe, above=2pt of being] {$\mathfrak{U}$: Understanding};
    \node [momentlabel=phasered, above=2pt of nothing] {$\mathfrak{A}$: Sublation};
    \node [momentlabel=exampleframe, below=2pt of becoming] {$\mathfrak{S}$: Speculation};

    \draw [-{Stealth[length=3mm]}, thick, bend left=20, color=phasered!80]
      (being) -- (nothing)
      node [midway, above, font=\tiny\itshape, text=phasered] {Self-sublation};

    \draw [-{Stealth[length=3mm]}, thick, bend right=15, color=exampleframe!80]
      (nothing) -- (becoming);

    \draw [-{Stealth[length=3mm]}, thick, bend left=15, color=promptframe!80]
      (being) -- (becoming)
      node [midway, left, font=\tiny\itshape, text=promptframe!80] {Preserve};

    \node at ($(being)!0.5!(nothing) + (0,-1.5cm)$) {\includegraphics[width=1.5cm]{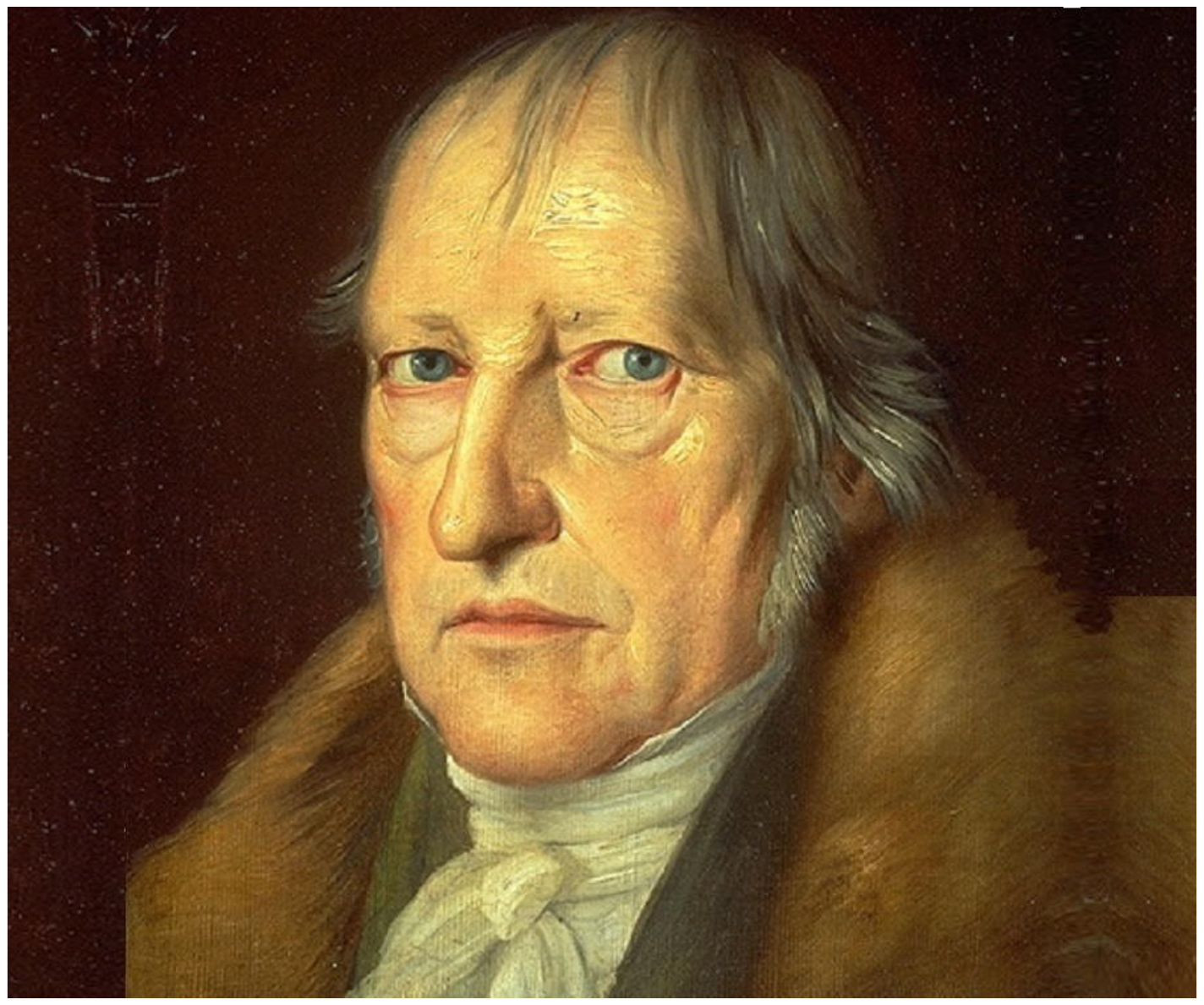}};

    \draw [-{Stealth[length=3mm]}, thick, dashed, bend right=25, draw=exampleframe!70]
      (becoming.west) -- ++(-2.2cm, 0) |- (being.west)
      node [midway, left, pos=0.2, font=\tiny\bfseries, text=exampleframe!80]
      {\rotatebox{90}{$T_{i+1} \gets S_i$}};
\end{tikzpicture}
}
\end{tcolorbox}
\vspace{-4pt}
\caption{\textbf{The Hegelian Dialectic.} A proposition $T_i$ ($\mathfrak{U}$) generates its opposition $A_i$ via self-sublation ($\mathfrak{A}$); their interaction produces a unified idea $T_{i+1}$ through speculation ($\mathfrak{S}$), which becomes the next proposition.}
\label{fig:dialectic}
\end{figure}

\paragraph{Step 1: Understanding}
The process begins with an initial proposition $T_0 \in \mathrm{Th}(\mathcal{L})$.

\paragraph{Step 2: Dialectical Self-sublation} At iteration $i$, $\mathcal{M}$ generates an opposition $A_i$ at temperature $\tau_A$:
\begin{defbox}
\tiny
\textbf{Sublation.}\; $A_i = \mathcal{M}(T_i, \tau_A, p_{\mathcal{A}})$ \;s.t.\; $\exists\, w\!:\; T_i \vdash w \,\wedge\, A_i \vdash \neg w$. The constant $\tau_A$ ensures iteration-invariant opposition.
\end{defbox}

\paragraph{Step 3: Speculation with Annealing}
$\mathcal{M}$ produces a unified response $S_i$:
\begin{defbox}
\tiny
\textbf{Speculation.}\; $S_i = \mathcal{M}(T_i, A_i, \tau(i), p_{\mathcal{S}})$\\[2pt]
s.t.\; $S_i \supseteq (T_i \cap A_i) \;\wedge\; S_i \not\vdash \bot$, where:
\begin{equation}
\label{eq:annealing}
\tau(i) = \tau_0 \cdot e^{-\theta i}, \quad \theta \geq 0,\; i \in \mathbb{N}
\end{equation}
Setting $\theta = 0$ recovers constant temperature $\tau(i) = \tau_0\;\forall\, i$.
\end{defbox}

\noindent The compositional dependency is:
\begin{align}
A_i &= \mathcal{A}_{\tau_A}(T_i),\; S_i = \mathcal{S}_{\tau(i)}(T_i, A_i) \notag\\
&\Longrightarrow\; S_i = \mathcal{S}_{\tau(i)}\!\bigl(T_i,\, \mathcal{A}_{\tau_A}(T_i)\bigr)
\end{align}
$\tau$ governs exploration-exploitation: high~$\tau$ promotes diverse syntheses, low~$\tau$ concentrates on refinement. The proposition is updated: $T_{i+1} \gets S_i$.

\paragraph{Probabilistic Formalization}
We formalize our complete framework as a joint probabilistic model. At each iteration~$t$, the joint distribution over the dialectical triple $(T_t, A_t, S_t)$ factorizes according to the three Hegelian moments:

\begin{propbox}
\vspace{-15pt}
\tiny
\begin{align}
\label{eq:joint-prob}
&P_{\tau(t), \tau_A}(S_t, A_t, T_t) = \notag\\
&\quad \underbrace{P(T_t)}_{\mathfrak{U}} \cdot \underbrace{P_{\tau_A}(A_t \mid T_t)}_{\mathfrak{A}} \cdot \underbrace{P_{\tau(t)}(S_t \mid T_t, A_t)}_{\mathfrak{S}}
\end{align}
where $\tau(t) \!\in\! \{\tau_0, \tau_0 e^{-\theta}, \ldots, \tau_0 e^{-\theta n}\}$ and $\tau_A \!\in\! \mathbb{R}^+$. The Markov property $T_{t+1} \!=\! S_t$ induces the chain:
\centering $T_0 \!\to\! (A_0, S_0) \!\to\! T_1 \!\to\! (A_1, S_1) \!\to\! \cdots \!\to\! T_N$
\end{propbox}

\subsection{Measuring Validity and Novelty}
Inspired by Minsky's \emph{``Society of Mind''}~\cite{minsky1988society,Zhuge2023MindstormsIN}, we employ $K$ LLMs to independently assess validity and novelty via majority voting. We adopt a greedy stopping heuristic: if any step fails MAMV, the process terminates. Algorithm~\ref{alg:LLM-self-reflection-llm-relevance-novelty} formalizes our self-reflection method.

\begin{defbox}
\tiny
\textbf{MAMV Decision Rule.}\; Given LLMs $\{M_1, \!\ldots\!, M_K\}$, prompts $p_V$, $p_N$, and triple $(T_i, A_i, S_i)$:
\begin{align}
v_k &= M_k(p_V, T_i, A_i, S_i) \in \{0,1\} \notag\\
n_k &= M_k(p_N, T_i, S_i) \in \{0,1\}
\end{align}
\vspace{-15pt}
\begin{align}
\mathrm{MAMV}(\cdot) &= \mathbb{1}\!\Bigl[\textstyle\sum_{k} v_k > \tfrac{K}{2}\Bigr] \wedge \mathbb{1}\!\Bigl[\textstyle\sum_{k} n_k > \tfrac{K}{2}\Bigr]
\end{align}
The \textbf{novelty score} is $\mathcal{N} = |\{i : \mathrm{MAMV}(\cdot) = 1\}| / N$.
\end{defbox}

\section{Prompt Engineering}
\label{sec:prompts}
Here we outline the prompts utilized in our method.
\subsection{Self-reflection Prompting}
\paragraph{Sublation:} 
Figures~\ref{fig:antithesis_idea} and~\ref{fig:antithesis_gsm8k} in Appendix~\ref{app:prompts} show the sublation prompts used for idea generation and math reasoning, respectively.
Engaging in self-debating is one way to generate diverse viewpoints. To simulate the sublation process, we employ an iterative self-debating, Solo Performance Prompting (SPP) strategy. SPP harnesses the model's theory of mind reasoning abilities by instructing it to ``split'' into various personas and collaborate on a given prompt through a brainstorming session among these personas~\cite{wang2024unleashingemergentcognitivesynergy}. Our framework requests the model to generate an arbitrary persona for each iteration of the self-dialectic.
The following instructs the model to practice SPP:
\begin{defbox}
\tiny
\textbf{Sublation Prompt.}\; \emph{Imagine you are someone X who has noticed \underline{a problem with or defect in a proposed view}. Produce \underline{an opposed view} on the same topic that corrects for the noted defect or problem.}
\end{defbox}
\noindent It is necessary to explicitly define the criteria for what constitutes potentially relevant types of defects. For ideation, we instruct the model with:
\begin{defbox}
\tiny
\textbf{Defect Categories.}\; Possible defects are:
\begin{enumerate} [noitemsep, topsep=2pt]
    \item The proposed view has constitutive elements that are not fully defined;
    \item The proposed view does not account for some relevant phenomena or examples (its content is incomplete);
    \item The proposed view is incompatible with some of the phenomena or examples that it is supposed to include;
    \item The proposed view cannot be fully defined on its own, but can be defined only in relation to another view.
\end{enumerate}
\end{defbox}

\paragraph{Speculation:}
Rather than choosing one view over the other, speculation seeks to unify opposing ideas by proposing a third perspective:
\begin{defbox}
\tiny
\textbf{Unification Instruction.}\; Produce a third view on the same topic that:
\begin{itemize}[noitemsep, topsep=2pt]
    \item Unifies the previous two views in relation to the defect or problem on which the two views were opposed to one another.
    \item This third view must be a \underline{unifying theory} that explains how \underline{the two views agree with} or are the same as one another and how \underline{the two views are opposed to} one another.
\end{itemize}
\end{defbox}
Figures~\ref{fig:synthesis_prompt_idea} and~\ref{fig:synthesis prompt_gsm8k} in Appendix~\ref{app:prompts} illustrate the speculation prompts used for idea generation and math reasoning.

\subsection{MAMV Prompting}
\paragraph{Validity:} To verify that the model has adhered to the speculation instructions, we provide MAMV with the instructions for the unifying process and ask each model to vote ``yes'' or ``no'' on whether the instructions are followed. Figure~\ref{fig:soundness} in Appendix~\ref{app:prompts} demonstrates this process.

\paragraph{Novelty:} Evaluating novelty is more challenging, as it requires domain experts who are familiar with all contributions in the field. Here, we define novelty as the introduction of new ideas that build upon the previous step's proposition i.e., $T_{i-1}$. The corresponding prompt is reported in Figure~\ref{fig:novelty} in Appendix~\ref{app:prompts}. We emphasize that our method generates new ideas that may not always be scientifically correct.
Our focus, however, is on the validity of the dialectical process provided through instructions for the generation of the unified idea, as well as the novelty of these ideas compared to the propositions and oppositions in previous steps. The scientific evaluation of these ideas could be examined by experts in the field using various scientific methods. We stress that evaluating novelty is not just about adding new information to the previous proposition; it also requires a thorough understanding of existing literature to ensure the ideas are truly unexplored. However, this does not undermine the effectiveness of our method as an early exploration but highlights the need for better evaluations, potentially involving human/AI experts to develop more effective ways.

\section{Experiments}
\label{case study}
 We apply our method to \emph{reasoning} and \emph{ideation} tasks, comparing constant against annealing temperature; the full setup and hyper-parameters (Table~\ref{table:params}) are in Appendix~\ref{app:expsetting}.

\begin{table*}[t]
\centering
\begin{minipage}[t]{0.56\textwidth}
\centering
\tiny
\setlength{\tabcolsep}{4pt}
\adjustbox{width=\linewidth}{%
\begin{tabular}{ccccccc}
\toprule
\rowcolor{promptframe!15}
\textbf{Model} & \textbf{Context} & \textbf{Prompt} & \textbf{GSM-8k} & \textbf{GSM-Symbolic} & \textbf{GSM-hard} & \textbf{MMLU Pro}\\
\midrule
&& Zero-shot & 0.863$\pm$0.004 & 0.758$\pm$0.006 & 0.597$\pm$0.003 & \underline{0.725$\pm$0.000}\\
&& Zero-shot+CoT & 0.863$\pm$0.004 & 0.760$\pm$0.004 & 0.587$\pm$0.004 & 0.714$\pm$0.000\\
\textbf{GPT-4o} &128k& Few-shot & 0.943$\pm$0.002 & 0.851$\pm$0.003 & 0.642$\pm$0.004 & 0.712$\pm$0.000\\
&& Few-shot+CoT & 0.944$\pm$0.002 & 0.843$\pm$0.005 & 0.640$\pm$0.008 & 0.723$\pm$0.000\\
&& Self-refine & \underline{0.954$\pm$0.004} & \underline{0.852$\pm$0.004} & \textbf{0.655$\pm$0.003} & \textbf{0.749$\pm$0.000}\\
\rowcolor{examplebg}&& \textbf{Dialectic (Ours)} & \textbf{0.955$\pm$0.000} & \textbf{0.884$\pm$0.004} & \underline{0.646$\pm$0.005} & \textbf{0.749$\pm$0.000}\\
\midrule
&& Zero-shot & 0.871$\pm$0.005 & 0.709$\pm$0.005 & 0.541$\pm$0.007 & 0.593$\pm$0.000\\
&& Zero-shot+CoT & 0.874$\pm$0.003 & 0.697$\pm$0.003 & 0.546$\pm$0.003 & \underline{0.639$\pm$0.000}\\
\textbf{GPT-4o-mini} &128k& Few-shot & 0.921$\pm$0.000 & 0.690$\pm$0.004 & 0.552$\pm$0.003 & 0.505$\pm$0.000\\
&& Few-shot+CoT & 0.921$\pm$0.001 & 0.708$\pm$0.009 & \underline{0.566$\pm$0.007} & 0.617$\pm$0.000\\
&& Self-refine & \underline{0.933$\pm$0.002} & \underline{0.709$\pm$0.003} & \textbf{0.574$\pm$0.005} & 0.627$\pm$0.000\\
\rowcolor{examplebg}&& \textbf{Dialectic (Ours)} & \textbf{0.940$\pm$0.003} & \textbf{0.770$\pm$0.008} & \textbf{0.574$\pm$0.006} & \textbf{0.673$\pm$0.000}\\
\midrule
&& Zero-shot & 0.828$\pm$0.000 & 0.513$\pm$0.000 & 0.491$\pm$0.000 & 0.422$\pm$0.000\\
&& Zero-shot+CoT & 0.816$\pm$0.000 & 0.516$\pm$0.000 & 0.501$\pm$0.000 & 0.454$\pm$0.000\\
\textbf{Qwen 2.5-7B-Instruct} &1M& Few-shot & 0.866$\pm$0.000 & 0.527$\pm$0.000 & \underline{0.517$\pm$0.000} & 0.411$\pm$0.000\\
&& Few-shot+CoT & 0.856$\pm$0.000 & 0.511$\pm$0.000 & 0.499$\pm$0.000 & 0.411$\pm$0.000\\
&& Self-refine & \underline{0.898$\pm$0.000} & \underline{0.587$\pm$0.000} & 0.515$\pm$0.000 & \underline{0.469$\pm$0.000}\\
\rowcolor{examplebg}&& \textbf{Dialectic (Ours)} & \textbf{0.907$\pm$0.000} & \textbf{0.623$\pm$0.000} & \textbf{0.548$\pm$0.000} & \textbf{0.499$\pm$0.000}\\
\midrule
&& Zero-shot & 0.869$\pm$0.000 & 0.741$\pm$0.000 & 0.560$\pm$0.000 & 0.695$\pm$0.000\\
&& Zero-shot+CoT & 0.861$\pm$0.000 & 0.742$\pm$0.000 & 0.559$\pm$0.000 & 0.706$\pm$0.000\\
\textbf{Phi-4} &16k& Few-shot & 0.949$\pm$0.000 & 0.825$\pm$0.000 & 0.619$\pm$0.000 & 0.687$\pm$0.000\\
&& Few-shot+CoT & 0.936$\pm$0.000 & 0.806$\pm$0.000 & 0.611$\pm$0.000 & 0.699$\pm$0.000\\
&& Self-refine & \underline{0.950$\pm$0.000} & \underline{0.839$\pm$0.000} & \underline{0.629$\pm$0.000} & \textbf{0.711$\pm$0.000}\\
\rowcolor{examplebg}&& \textbf{Dialectic (Ours)} & \textbf{0.954$\pm$0.000} & \textbf{0.849$\pm$0.000} & \textbf{0.635$\pm$0.000} & \underline{0.709$\pm$0.000}\\
\midrule
&& Zero-shot & 0.801$\pm$0.000 & 0.669$\pm$0.000 & 0.454$\pm$0.000 & 0.542$\pm$0.000\\
&& Zero-shot+CoT & 0.854$\pm$0.000 & 0.626$\pm$0.000 & 0.506$\pm$0.000 & 0.551$\pm$0.000\\
\textbf{Qwen3-8B} &32k& Few-shot & 0.916$\pm$0.000 & 0.658$\pm$0.000 & 0.578$\pm$0.000 & 0.565$\pm$0.000\\
&& Few-shot+CoT & \underline{0.921$\pm$0.000} & 0.723$\pm$0.000 & 0.578$\pm$0.000 & 0.572$\pm$0.000\\
&& Self-refine & \textbf{0.933$\pm$0.000} & \underline{0.774$\pm$0.000} & 0.\underline{585$\pm$0.000} & \underline{0.601$\pm$0.000}\\
\rowcolor{examplebg}&& \textbf{Dialectic (Ours)} & \textbf{0.933$\pm$0.000} & \textbf{0.784$\pm$0.000} & \textbf{0.658$\pm$0.000} & \textbf{0.613$\pm$0.000}\\
\bottomrule
\end{tabular}}
\caption{Dialectical self-reflection vs.\ other prompting. The \colorbox{examplebg}{\small highlighted} \textbf{Dialectic (Ours)} row reports the stronger of our constant and annealing schedules per benchmark. \textbf{Bold} = best, \underline{underline} = second best in each column.}
\label{tbl:main_results}
\end{minipage}\hfill
\begin{minipage}[t]{0.42\textwidth}
\centering
\tiny
\setlength{\tabcolsep}{5pt}
\renewcommand{\arraystretch}{2.8}
\adjustbox{width=\linewidth}{%
\begin{tabular}{llcccc}
\toprule
\rowcolor{promptframe!15}
\textbf{Model} & \textbf{Temperature} & \textbf{GSM-8k} & \textbf{GSM-Symbolic} & \textbf{GSM-hard} & \textbf{MMLU Pro}\\
\midrule
\multirow{2}{*}{\textbf{GPT-4o}}
 & Constant  & 0.953$\pm$0.004 & 0.879$\pm$0.003 & \textbf{0.646$\pm$0.005} & \textbf{0.749$\pm$0.000}\\
 & Annealing & \textbf{0.955$\pm$0.000} & \textbf{0.884$\pm$0.004} & 0.644$\pm$0.006 & \textbf{0.749$\pm$0.000}\\
\midrule
\multirow{2}{*}{\textbf{GPT-4o-mini}}
 & Constant  & 0.939$\pm$0.004 & \textbf{0.770$\pm$0.008} & 0.572$\pm$0.005 & \textbf{0.673$\pm$0.000}\\
 & Annealing & \textbf{0.940$\pm$0.003} & 0.769$\pm$0.000 & \textbf{0.574$\pm$0.006} & 0.663$\pm$0.000\\
\midrule
\multirow{2}{*}{\textbf{Qwen 2.5-7B-Instruct}}
 & Constant  & \textbf{0.907$\pm$0.000} & 0.615$\pm$0.000 & \textbf{0.548$\pm$0.000} & \textbf{0.499$\pm$0.000}\\
 & Annealing & 0.895$\pm$0.000 & \textbf{0.623$\pm$0.000} & 0.547$\pm$0.000 & \textbf{0.499$\pm$0.000}\\
\midrule
\multirow{2}{*}{\textbf{Phi-4}}
 & Constant  & 0.951$\pm$0.000 & \textbf{0.849$\pm$0.000} & 0.633$\pm$0.000 & 0.707$\pm$0.000\\
 & Annealing & \textbf{0.954$\pm$0.000} & 0.844$\pm$0.000 & \textbf{0.635$\pm$0.000} & \textbf{0.709$\pm$0.000}\\
\midrule
\multirow{2}{*}{\textbf{Qwen3-8B}}
 & Constant  & \textbf{0.933$\pm$0.000} & \textbf{0.784$\pm$0.000} & \textbf{0.658$\pm$0.000} & 0.605$\pm$0.000\\
 & Annealing & 0.932$\pm$0.000 & 0.780$\pm$0.000 & 0.645$\pm$0.000 & \textbf{0.613$\pm$0.000}\\
\bottomrule
\end{tabular}}
\caption{\textbf{Ablation study}: Accuracy for constant vs.\ annealing speculation temperature.}
\label{tbl:ablation}
\end{minipage}
\end{table*}

\subsection{Quantitative Results: Reasoning}
Table~\ref{tbl:main_results} compares our method against standard prompting and multi-turn Self-refine, and Table~\ref{tbl:ablation} isolates the temperature schedule.
\achead{Comparison with baselines.}
\emph{\textcolor{violet}{Our method outperforms all prompting baselines, including Self-refine, on GSM-Symbolic for all five models, and is best or tied on GSM-8k}}. The \emph{relative gains} over the best baseline on GSM-Symbolic are substantial: GPT-4o $0.852\!\to\!0.884$ ($+3.8\%$), GPT-4o-mini $0.709\!\to\!0.770$ ($+8.6\%$), and Qwen~2.5-7B $0.587\!\to\!0.623$ ($+6.1\%$). On the near-saturated GSM-8k (baselines already exceed $0.94$ for the GPT models) we top every model, tying the strong Self-refine baseline ($0.933$) on Qwen3-8B. GSM-hard and MMLU Pro are more competitive, yet the reasoning-tuned Qwen3-8B posts the largest single gain of any model on GSM-hard ($0.585\!\to\!0.658$, \emph{\textcolor{violet}{$+12.5\%$}}). Overall we top GSM-hard everywhere except GPT-4o, and on MMLU Pro we lead on GPT-4o-mini and both Qwen models, tie GPT-4o, and trail Self-refine by just $0.002$ on Phi-4.

\begin{algorithm}[!ht]
\caption{Hegelian Dialectical Self-Reflection}
\label{alg:LLM-self-reflection-llm-relevance-novelty}
\begin{tcolorbox}[
  enhanced,
  colback=white, colframe=promptframe,
  boxrule=0.7pt, arc=3pt,
  shadow={1pt}{-1pt}{0pt}{black!10},
  left=4pt, right=4pt, top=4pt, bottom=4pt,
]
\tiny
\setlength{\baselineskip}{10pt}
\textbf{\color{promptframe}Part 1:} Hegelian Dialectical Self-Reflection
\vspace{3pt}
\hrule
\vspace{5pt}
\begin{algorithmic}[1]
\STATE \textbf{Input:} LLM $\mathcal{M}$, prompts $p_{\mathcal{A}}$, $p_{\mathcal{S}}$, proposition $T_0$,
\STATE \hspace{2.6em} temperatures $\tau_0$, $\tau_A$, decay $\theta$, max iterations $N$
\STATE \textbf{Output:} Final unified idea $S^*$
\vspace{3pt}
\STATE $T \gets T_0$ \COMMENT{\color{exampleframe}Initialize proposition}
\vspace{2pt}
\FOR{$i = 0$ \textbf{to} $N-1$}
    \vspace{2pt}
    \STATE \colorbox{promptframe!8}{$A_i \gets \mathcal{M}(T_i,\; \tau_A,\; p_{\mathcal{A}})$} \COMMENT{\color{phasered}$\mathfrak{A}$: Sublation}
    \vspace{2pt}
    \STATE $\tau(i) \gets \tau_0 \cdot e^{-\theta i}$ \COMMENT{\color{exampleframe}Update temperature}
    \vspace{2pt}
    \STATE \colorbox{examplebg}{$S_i \gets \mathcal{M}(T_i,\; A_i,\; \tau(i),\; p_{\mathcal{S}})$} \COMMENT{\color{exampleframe}$\mathfrak{S}$: Speculation}
    \vspace{2pt}
    \IF{$\mathrm{MAMV}(T_i, A_i, S_i) = 1$}
        \STATE $T_{i+1} \gets S_i$ \COMMENT{\color{exampleframe}Accept \& advance}
    \ELSE
        \STATE $S^* \gets T_i$; \textbf{break} \COMMENT{\color{phasered}Halt}
    \ENDIF
    \vspace{2pt}
\ENDFOR
\vspace{2pt}
\STATE \textbf{return} $S^* \gets S_{i}$
\end{algorithmic}

\vspace{6pt}
\hrule\vspace{1pt}\hrule
\vspace{6pt}

\textbf{\color{promptframe}Part 2:} Multi-Agent Majority Voting (MAMV)\label{alg:MAMV}
\vspace{3pt}
\hrule
\vspace{5pt}
\begin{algorithmic}[1]
\STATE \textbf{Input:} Triple $(T, A, S)$, prompts $p_V$, $p_N$,
\STATE \hspace{2.6em} LLM panel $\{M_1, \ldots, M_K\}$
\STATE \textbf{Output:} Decision $d \in \{0, 1\}$
\vspace{3pt}
\FOR{$k = 1$ \textbf{to} $K$}
    \vspace{2pt}
    \STATE \colorbox{promptframe!8}{$v_k \gets M_k(p_V,\; T,\; A,\; S)$} \COMMENT{\color{phasered}Validity}
    \vspace{2pt}
    \STATE \colorbox{examplebg}{$n_k \gets M_k(p_N,\; T,\; S)$} \COMMENT{\color{exampleframe}Novelty}
    \vspace{2pt}
\ENDFOR
\vspace{3pt}
\STATE $d \gets \mathbb{1}\!\bigl[\sum_{k} v_k > \tfrac{K}{2}\bigr] \;\wedge\; \mathbb{1}\!\bigl[\sum_{k} n_k > \tfrac{K}{2}\bigr]$
\vspace{2pt}
\STATE \textbf{return} $d$
\end{algorithmic}
\end{tcolorbox}
\end{algorithm}
\achead{Task-specific analysis.}
GSM-8k/hard target multi-step arithmetic, while Symbolic requires abstract symbolic manipulation. Gains are \emph{\textcolor{violet}{most pronounced on the symbolic and knowledge tasks}}: the sublation-speculation cycle surfaces and repairs the structural errors that dominate rule-based reasoning and that a single forward pass leaves uncorrected. Qwen3-8B is the exception: already strong on symbolic structure, its gains concentrate on GSM-hard, where iterative error-correction pays off most on longer arithmetic chains.
\achead{Annealing vs.\ constant temperature.}
Table~\ref{tbl:ablation} compares the two schedules (Eq.~\ref{eq:annealing}). Neither dominates: annealing wins GSM-8k for the GPT and Phi-4 models (three of five), while both open-weight Qwen models (2.5-7B and Qwen3-8B) prefer a constant temperature on GSM-8k and GSM-hard; elsewhere the schedules trade off and stay within error bars on most MMLU~Pro cells. 
\achead{Variance reduction.}
\emph{\textcolor{violet}{Beyond accuracy, our method consistently reduces output variance}}. By Proposition~\ref{prop:speculation}, the preservation constraint ($S_i \supseteq T_i \cap A_i$) retains validated premises, progressively narrowing the solution space.
\achead{Effect of context length.}
 Dialectical prompting is multi-turn: each iteration appends the full proposition-opposition-unification history. Larger windows hold the whole trajectory without truncation, keeping the reasoning chain intact, so gains grow with context; smaller windows (e.g., Phi-4, 16k) compress it, giving more modest yet still best-among-all gains. \emph{\textcolor{violet}{The effect saturates once the window fits the trajectory}}: our three-iteration runs need only a few thousand tokens, so beyond that the base model's reasoning matters more than window size. Qwen3-8B shows this: with just 32k, below GPT-4o's 128k or Qwen~2.5's 1M, its reasoning-tuned backbone gives the largest GSM-hard gain of any model (Table~\ref{tbl:main_results}), while Phi-4's 16k is where context truly binds. Figure~\ref{fig:example_gsm} (Appendix~\ref{app:iters}) shows a complete dialectical iteration.

\subsection{Qualitative Results: Scientific Ideation}
\begin{figure}[H]
\centering
\begin{subfigure}{\linewidth}
\centering
\begin{tcolorbox}[enhanced, colback=white, colframe=promptframe, boxrule=0.8pt, arc=3pt, shadow={1.5pt}{-1.5pt}{0pt}{black!15}, left=2pt, right=2pt, top=2pt, bottom=2pt]
\includegraphics[width=\linewidth]{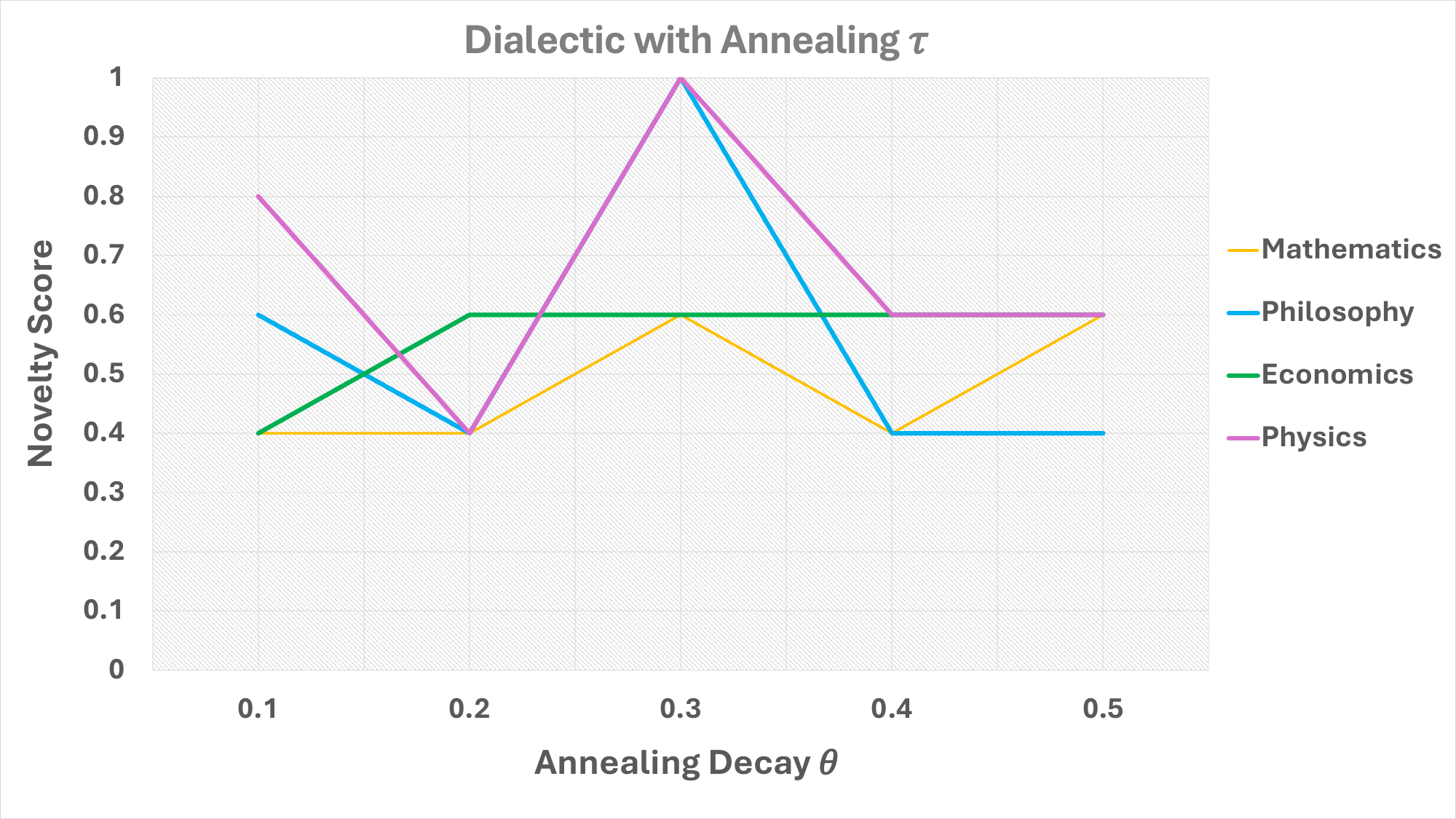}
\end{tcolorbox}
\caption{\textbf{Annealing decay $\theta$.} Low decay ($\theta \!\to\! 0$) mimics constant high temperature; high decay ($\theta \!\to\! 1$) mimics low temperature. The optimal $\theta = 0.3$ maximizes $\mathcal{N}$.}
\label{fig:annealing}
\end{subfigure}

\vspace{0.7em}

\begin{subfigure}{\linewidth}
\centering
\begin{tcolorbox}[enhanced, colback=white, colframe=promptframe, boxrule=0.8pt, arc=3pt, shadow={1.5pt}{-1.5pt}{0pt}{black!15}, left=2pt, right=2pt, top=2pt, bottom=2pt]
\includegraphics[width=\linewidth]{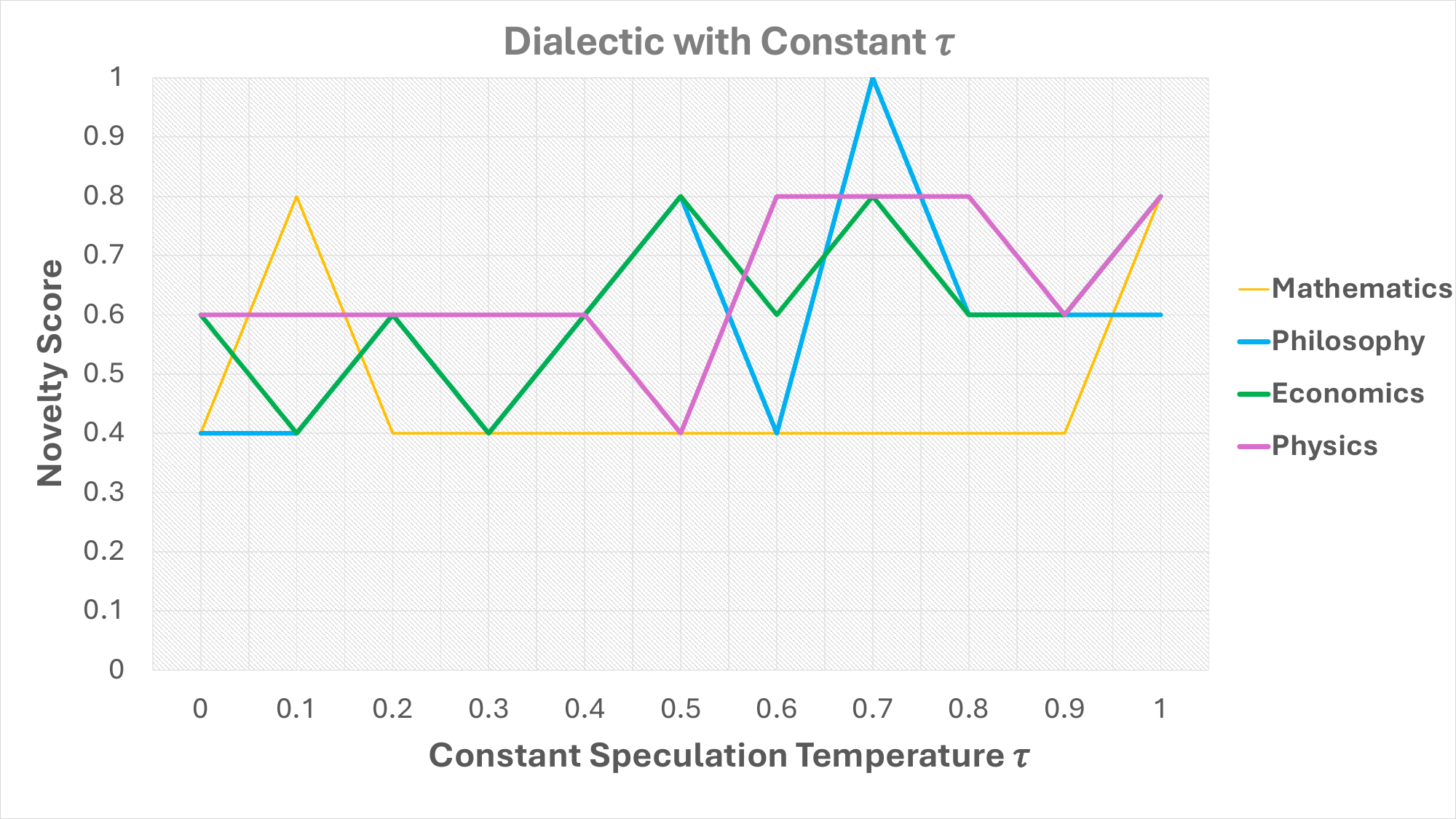}
\end{tcolorbox}
\caption{\textbf{Constant temperature $\tau$.} Higher $\tau$ increases novelty but risks topic drift; lower $\tau$ maintains coherence at the cost of reduced creativity.}
\label{fig:abblation}
\end{subfigure}
\caption{Effect of speculation temperature on novelty $\mathcal{N}$: (a)~dynamic annealing schedule and (b)~constant temperature.}
\label{fig:temp_configs}
\end{figure}

 \paragraph{Observations and Key Findings}
Across hundreds of GPT-4o runs spanning annealing and constant $\tau$ and slightly varied prompts, we observe:\\
\textbf{1)} Identical configs and prompts can still yield varying responses, with different step counts and novelty scores, though ideas from one configuration usually stay on the same topic.\\
\textbf{2)} A high constant speculation $\tau$ can yield either out-of-topic or genuinely novel unified ideas. \\
 \textbf{3)} 
 The most novel unifications occur early; over iterations, ideas grow more refined and absorb opposing viewpoints, making meaningful opposition harder to generate and sometimes repeating it, a sign it has become comprehensive and stable.\\
\textbf{4)} For reasoning, the unified idea's level of agreement and the opposition's intensity (opposing qualifications vs.\ fundamental premises) act as prompt-based hyper-parameters shaping the self-debate.\\
\textbf{5)} Both temperatures shape novelty: for sublation, low $\tau_A$ yields arguments that directly address the proposition, while high $\tau_A$ yields near-orthogonal, contradictory perspectives whose resolution qualifies both sides into self-consistent but too-agreeable discussions ($\tau_A = 0.5$ is a compromise). For speculation (Figure~\ref{fig:annealing}), low decay ($\theta \to 0$) mimics a constant high temperature while high decay over-suppresses creativity, with $\theta = 0.3$ maximizing novelty; under constant $\tau$, higher values raise novelty but risk topic drift (Figure~\ref{fig:abblation}).\\
\textbf{6)} Our hyper-parameters and prompts are not universal and must be tuned per topic.\\
\textbf{7)} More statements in the proposition raise the likelihood of opposing views, yielding more meaningful opposition and additional steps. A full dialectical iteration, as well as ideation examples for math, physics, economics, and philosophy, appear in Appendices~\ref{app:iters} and~\ref{app:results}.


\input{limitations_appendix}

\section{Concluding Remarks}
\label{Concluding Remarks}
We framed LLM self-reflection as a self-dialectical process grounded in Hegel's dialectics: a cycle of understanding, sublation, and speculation that challenges an initial proposition with opposing views and refines them into a unified idea combining the strengths of both. We studied how speculation temperature affects novelty under constant and dynamic annealing schedules, assessing novelty with MAMV, a multi-agent collaborate-and-vote system. Our method yields significant gains on mathematical, symbolic, and knowledge-intensive reasoning (GSM-8k, GSM-hard, GSM-Symbolic, and MMLU Pro) and promising idea generation: high constant temperatures occasionally produced off-topic or highly novel ideas, while annealing surfaced the most innovative ideas early, growing more nuanced and stable over time.

\newpage
\bibliography{main}
\onecolumn
\section*{Appendix}
\appendix
\input{experimental_setting}

\section{Prompt Templates}
\label{app:prompts}
The full prompt templates referenced in Section~\ref{sec:prompts} are shown below (Figures~\ref{fig:antithesis_idea}--\ref{fig:novelty}).
\input{appendix_prompt}
\input{appendix_examples}
\end{document}

%% file: limitations_appendix.tex
\section{Limitations and Future Directions}
\label{app:limits}
In this section, we briefly discuss some limitations of our current framework and experimental setup.
\paragraph{Deliberate adaptation and interpretation choices in Hegelian dialectic}We formulate Hegel's dialectic as a 2-step progress, where the flow from proposition to opposition and then unified idea does not follow any sort of logical necessity as proposed by~\cite{ sep-hegel-dialectics, Kaufmann1965-KAUH, inoue2014dialectical}. This also means that the perspective from which the opposition introduces its opposition is arbitrary, i.e. delegated to the LLM agent. (2) The unified idea always maintains a self-consistent position. In Hegel’s dialectic, statements can remain ambiguous, where they are true from one perspective and false from the other, without requiring a resolution~\cite{ hegel_phenomenology_of_spirit, sep-hegel-dialectics}. Interested readers can refer to~\cite{ sep-hegel-dialectics, inoue2014dialectical} for more details.
\paragraph{Measuring novelty of a statement is challenging} as it involves subjective assessment and context-awareness. Novelty is not just about adding new information, it also requires understanding existing literature to ensure that the idea has not been previously explored. Comparing an idea solely based on an initial proposition does not account for depth of existing research.
Our novelty prompt has inherent limitations. It does not explicitly define whether altering the strength of an argument is considered novel, even if all points have been previously mentioned. Also, it remains unclear whether the model should regard the negation of existing premises as novel, as it involves manipulating existing information rather than introducing new information. However, we view such negation as novel.
\paragraph{Difficulties in distinguishing creativity from LLM remembering in baseline evaluation} Evaluating new ideas against pre-proposed unified idea is extremely challenging, as distinguishing whether a model is merely recalling training information or generating creative ideas is challenging.
\paragraph{Measuring the effect of annealing due to randomness of generation:} We often observed that, a constant $\tau$ either fails to produce novel content when set too low or leads to irrelevant information when set too high. However, due to inherent randomness in the generation, it is not straightforward to generalize this observation. Conducting statistical significance tests over multiple rounds of prompting can help to evaluate this observation more effectively. 
\paragraph{Reproducibility of results:} The inherent randomness in the generation, coupled with constantly evolving nature of LLMs and the lack of control, especially when using black-box models (e.g. GPT-x), makes it difficult to reproduce the results which is essential in scientific settings. While white-box models offer better control over outputs, they may not be as powerful as black-box competitors. 
\paragraph{Repetition and randomness in API calls}
We have occasionally observed identical opposition and unification paragraphs, as well as main points, raising concerns about the extent of randomness and the possible presence of hidden caching mechanisms during these experiments.
\paragraph{Need for domain expert LLMs} Utilizing the extensive knowledge of domain-specific LLM experts can help identify unique contributions and ensure comprehensive coverage of the field, making it more feasible to measure novelty.

\paragraph{Future directions.} In this paper, for simplicity, we used a constant $\tau_A$. However, as we discussed earlier, it directly affects novelty, with lower $\tau_A$ producing arguments directly addressing the initial proposition and higher $\tau_A$ results in orthogonal perspectives that lead to epistemological inquiries. Investigating multiple oppositions with varying $\tau_A$ simultaneously, generating unified ideas, and backtracking from undesired outcomes is worthwhile. We reserve this exploration for future research.
\par Currently, with the novelty score stop condition, our framework strives to resolve dialectical contradictions every cycle. In future implementations, the statements from opposing view points might be explicitly considered as ambiguous and not integrated into the unified idea until enough iterations have passed to resolve them.
\par Meanwhile, in the MAD setting, the Multi-Persona framework addresses the degeneracy-of-thought issue in naive self-reflection by pre-assigning one agent to express viewpoints and another to oppose them~\cite{liang2024encouragingdivergentthinkinglarge}. By adjusting the likelihood of agreement in debate protocols, the Multi-Persona MAD framework surpasses other MAD frameworks in the Q\&A setting by tuning the agreeability of debating agents.~\cite{liang2024encouragingdivergentthinkinglarge, smit2024goingmadlookmultiagent}. Similarly, the decisiveness of the speculation step can be tuned and compared with the performance of agreeable/disagreeable MAD frameworks.

%% file: experimental_setting.tex
\section{Experimental Setting}
\label{app:expsetting}
\paragraph{Models.} We utilize GPT family models for both quantitative and qualitative experiments. For idea generation, we employ \textit{GPT-4o mini}, \textit{GPT4-32k}, and \textit{GPT-4} within the MAMV framework, while \textit{GPT-4o} serves as the core model for dialectical self-reflection. For reasoning, we additionally evaluate on \textit{Qwen~2.5-7B} and \textit{Phi-4}.
\paragraph{Dialectical Iterations and MAMV.} We set a maximum iteration count $N$ per experiment. For idea generation, an external MAMV module evaluates the validity and novelty of each unified idea; the process halts if either criterion fails. For reasoning tasks, we do not use MAMV, isolating the effect of dialectical self-reflection from multi-agent voting.
\paragraph{Sublation Temperature $\tau_A$.} During the sublation step, $\tau_A$ directly influences novelty: lower $\tau_A$ produces arguments closely engaging with the proposition, while higher $\tau_A$ generates more divergent perspectives. We set $\tau_A = 0.5$ as a compromise.
\paragraph{Speculation Temperature.} We conduct two sets of experiments: one with the annealing schedule $\tau(i) = \tau_0 e^{-\theta i}$ and one with constant temperature ($\theta = 0$). We grid-searched $\theta$ and chose a high $\tau_0$ to ensure initial creativity. Figures~\ref{fig:annealing} and~\ref{fig:abblation} present the optimal $\theta$ and constant $\tau$ respectively.
\paragraph{Hyper-parameters.} Tables~\ref{table:params} and~\ref{table:temp} detail our experimental hyper-parameters.

\begin{table}[h!]
\centering
\scriptsize
\begin{tabular}{l c}
  \toprule
  \rowcolor{promptframe!15}
  \textbf{Hyper-parameter} & \textbf{Value} \\
  \midrule
  Initial temperature $\tau_0$ & 0.7 \\
  Constant temperature for opposition $\tau_A$ & 0.5 \\
  Decay constant $\theta$ & $[0.1,\, 0.5]$ \\
  Max iterations (idea generation) $N$ & 5 \\
  Max iterations (math reasoning) $N$ & 3 \\
  \bottomrule
\end{tabular}
\caption{Experimental hyper-parameters.}
\label{table:params}
\end{table}

\begin{table}[h!]
\centering
\scriptsize
\begin{tabular}{llcccc}
\toprule
\rowcolor{promptframe!15}
\textbf{Dataset} & \textbf{Model} & \multicolumn{2}{c}{\textbf{Constant}} & \multicolumn{2}{c}{\textbf{Annealing}} \\
\cmidrule(lr){3-4} \cmidrule(lr){5-6}
\rowcolor{promptframe!15}
& & $\tau_0$ & $\tau_A$ & $\tau_0$ & $\tau_A$ \\
\midrule

\multirow{6}{*}{GSM-Hard}
& GPT-4o                 & 0.7 & 0.5 & 0.7 & 0.5 \\
& GPT-4o-mini            & 0.7 & 0.5 & 0.7 & 0.5 \\
& GPT4-32k               & 1.0 & 0.3 & 1.0 & 0.3 \\
& Phi-4                  & 1.0 & 0.5 & 1.0 & 0.5 \\
& Qwen2.5                & 1.0 & 0.3 & 1.0 & 0.3 \\
& Phi-4-mini-instruct    & 1.0 & 0.5 & 1.0 & 0.5 \\

\midrule

\multirow{6}{*}{GSM-Symbolic}
& GPT-4o                 & 0.7 & 0.5 & 0.7 & 0.5 \\
& GPT-4o-mini            & 0.7 & 0.5 & 0.7 & 0.5 \\
& GPT4-32k               & 0.7 & 0.5 & 0.7 & 0.5 \\
& Phi-4                  & 1.0 & 0.5 & 1.0 & 0.5 \\
& Qwen2.5                & 1.0 & 0.3 & 1.0 & 0.3 \\
& Phi-4-mini-instruct    & 1.0 & 0.5 & 1.0 & 0.5 \\

\midrule

\multirow{6}{*}{GSM8k}
& GPT-4o                 & 0.7 & 0.5 & 0.7 & 0.5 \\
& GPT-4o-mini            & 0.7 & 0.5 & 0.7 & 0.5 \\
& GPT4-32k               & 1.0 & 0.3 & 1.0 & 1.0 \\
& Phi-4                  & 0.7 & 0.5 & 0.7 & 0.5 \\
& Qwen2.5                & 1.0 & 0.3 & 0.7 & 1.0 \\
& Phi-4-mini-instruct    & 1.0 & 0.5 & 1.0 & 0.5 \\

\bottomrule
\end{tabular}
\caption{Comparison of $\tau_0$ and $\tau_A$ values across datasets and models.}
\label{table:temp}
\end{table}

%% file: appendix_prompt.tex
\begin{figure}[H]
\begin{tcolorbox}[enhanced, colback=promptframe!3, colframe=promptframe, boxrule=0.7pt, arc=3pt, shadow={1pt}{-1pt}{0pt}{black!10}, left=3pt, right=3pt, top=3pt, bottom=3pt, title={\small\bfseries Sublation Prompt $(p_{\mathcal{A}})$ -- Qualitative Experiment}, coltitle=white]
\scriptsize

\begin{iterbox}{Instructions}
You are following Hegel's dialectical method.
You have access to a Thesis.

\textbf{Your task:}
\begin{enumerate}[leftmargin=*, nosep]
    \item Read the Thesis below (and any prior context if needed).
    \item Imagine you are someone X, who has an orthogonal OR diametrically opposed perspective on the same topic.
    \item Produce an Antithesis that is self-consistent, follows an argumentative structure, and contradicts the Thesis.
    \item If the Thesis already qualifies all anticipated counterarguments, the Antithesis should be a new competing thesis with the same goal as the current Thesis.
    \item Include novel, opinionated perspectives or angles beyond what has been stated so far.
    \item Provide a concise ``Summary of Antithesis'' at the end.
\end{enumerate}
\end{iterbox}

\begin{iterbox}{Response Format}
\textcolor{phasered}{\textbf{State who you are and your perspective:}}\\
X: [Name of the person with the Antithesis perspective]\\[2pt]

\textcolor{phasered}{\textbf{Premises:}}
\begin{itemize}[nosep]
    \item Premise 1
    \item Premise 2
    \item ($\cdots$ as many premises as needed)
\end{itemize}

\textcolor{phasered}{\textbf{Reasoning:}} Explain how these premises contradict or challenge the Thesis, adding original, opinionated perspectives.\\[2pt]
\textcolor{phasered}{\textbf{Conclusion:}} The final claim that directly opposes or contradicts the Thesis.\\[2pt]
\textcolor{exampleframe}{\textbf{Antithesis:}} Detailed explanation.\\[2pt]
\textcolor{exampleframe}{\textbf{Summary of Antithesis:}} One- or two-sentence summary capturing the essence of this Antithesis.
\end{iterbox}
\end{tcolorbox}
\caption{Prompt used during the sublation process in the idea generation experiment.}
\label{fig:antithesis_idea}
\end{figure}

\begin{figure}[H]
\begin{tcolorbox}[enhanced, colback=promptframe!3, colframe=promptframe, boxrule=0.7pt, arc=3pt, shadow={1pt}{-1pt}{0pt}{black!10}, left=3pt, right=3pt, top=3pt, bottom=3pt, title={\small\bfseries Speculation Prompt $(p_{\mathcal{S}})$ -- Qualitative Experiment}, coltitle=white]
\scriptsize

\begin{iterbox}{Instructions}
You are following Hegel's dialectical method.
You have access to the summary of the previous conversations, including the Thesis and Antithesis.

\textbf{Your task:}
\begin{enumerate}[leftmargin=*, nosep]
    \item Read the Thesis and Antithesis below.
    \item Produce a Synthesis that either:
    \begin{itemize}[nosep]
        \item Qualifies the contradictory statements so that Thesis and Antithesis unify without contradiction, OR
        \item Negates contradictory statements, explicitly listing which ones are negated and explaining why.
    \end{itemize}
    \item Present a coherent, integrated perspective that resolves or clarifies these contradictions.
    \item Provide a short ``Summary of Synthesis'' statement, which will be used as the next Thesis.
\end{enumerate}
\end{iterbox}

\begin{iterbox}{Response Format}
\textcolor{phasered}{\textbf{Thesis UNION Antithesis:}}\\[2pt]

\textcolor{phasered}{\textbf{Premises:}}
\begin{itemize}[nosep]
    \item Premise 1
    \item Premise 2
    \item ($\cdots$ as many premises as needed)
\end{itemize}

\textcolor{phasered}{\textbf{Reasoning:}}\\[2pt]
\textcolor{phasered}{\textbf{Conclusion:}}\\[2pt]

\textcolor{exampleframe}{\textbf{Synthesis:}} Detailed explanation of how contradictions are integrated or negated. If qualifying contradictions, detail how they are harmonized. If negating specific statements, list them and briefly explain why. If both are lacking an important premise, introduce that confounding or missing premise. Explain how this new perspective leads to a coherent, possibly novel position.\\[2pt]

\textcolor{exampleframe}{\textbf{Summary of Synthesis (Next Thesis):}} One- or two-sentence statement that unifies the positions and serves as the next Thesis.
\end{iterbox}
\end{tcolorbox}
\caption{Prompt used during the speculation process in the idea generation experiment.}
\label{fig:synthesis_prompt_idea}
\end{figure}

\begin{figure}[H]
\begin{tcolorbox}[enhanced, colback=promptframe!3, colframe=promptframe, boxrule=0.7pt, arc=3pt, shadow={1pt}{-1pt}{0pt}{black!10}, left=3pt, right=3pt, top=3pt, bottom=3pt, title={\small\bfseries Sublation Prompt $(p_{\mathcal{A}})$ -- Quantitative Experiment (Reasoning)}, coltitle=white]
\scriptsize

\begin{iterbox}{Instructions}
You are following Hegel's dialectical method.

You have access to the question: \{question\} and the following proposed solution: \{solution 1\}.

\textbf{Your task:}
\begin{enumerate}[leftmargin=*, nosep]
    \item Read the proposed solution carefully. If you notice any problem or defect, proceed to task 2; otherwise return the proposed solution.
    \item Imagine you are someone X who has noticed a problem with or defect in the proposed solution. If and only if the proposed solution is not correct, produce an opposed solution that corrects the noted defect.
    Possible defects include, but are not limited to:
    \begin{itemize}[nosep]
        \item \textbf{Arithmetic errors:} incorrect calculation, order of operations, decimal mismanagement, fraction handling.
        \item \textbf{Misinterpretation:} failure to extract key information, incorrect units/conversions, incorrect assumptions.
        \item \textbf{Logical errors:} flawed reasoning steps, skipping key intermediate steps.
        \item \textbf{Over-complication:} introducing unnecessary complexity.
        \item \textbf{Formatting errors:} incomplete solution, inconsistent format, ambiguity.
        \item \textbf{Conceptual misunderstanding:} misunderstanding the mathematical concept, confusing problem types.
    \end{itemize}
\end{enumerate}

For example: \{Few-shot examples\}\\
Let's think step by step.
\end{iterbox}

\begin{iterbox}{Response Format}
\textcolor{phasered}{\textbf{Defects or problems:}}
\begin{enumerate}[nosep]
    \item Defect/Problem 1
    \item Defect/Problem 2
    \item (\ldots\ as many defects as needed)
\end{enumerate}
Q: \{question\}\\
A:\\
Ensure the answer~A concludes with ``The answer is\ldots''
\end{iterbox}
\end{tcolorbox}
\caption{Prompt used during the sublation process in the GSM8k evaluation.}
\label{fig:antithesis_gsm8k}
\end{figure}
\begin{figure}[H]
\begin{tcolorbox}[enhanced, colback=promptframe!3, colframe=promptframe, boxrule=0.7pt, arc=3pt, shadow={1pt}{-1pt}{0pt}{black!10}, left=3pt, right=3pt, top=3pt, bottom=3pt, title={\small\bfseries Speculation Prompt $(p_{\mathcal{S}})$ -- Quantitative Experiment (Reasoning)}, coltitle=white]
\scriptsize

\begin{iterbox}{Instructions}
You are following Hegel's dialectical method.\\
You have access to two opposing solutions for a given math question.

\textbf{Your task:}
\begin{itemize}[leftmargin=*, nosep]
    \item Read the given opposing solutions below.
    \item Produce a third solution for the same problem such that:
    \begin{itemize}[nosep]
        \item It unifies the previous two solutions in relation to the defects or problems that made them opposed.
        \item This third solution must capture how the two solutions agree and resolve the reported defects.
    \end{itemize}
\end{itemize}
\end{iterbox}

\begin{iterbox}{Response Format}
\textcolor{phasered}{\textbf{Agreements}} between solution 1 and solution 2:
\begin{enumerate}[nosep]
    \item Point 1 \quad\ldots
\end{enumerate}
\textcolor{phasered}{\textbf{Defects}} with solution 1:
\begin{enumerate}[nosep]
    \item Defect/Problem 1 \quad\ldots
\end{enumerate}
\textcolor{phasered}{\textbf{Defects}} with solution 2:
\begin{enumerate}[nosep]
    \item Defect/Problem 1 \quad\ldots
\end{enumerate}
Solution 1: `\{solution 1\}'\\
Solution 2: `\{solution 2\}'\\
Q: \{question\}\\
A:\\
Ensure the answer is the same as the unified solution but concludes with ``The answer is\ldots''\\
Let's think step by step.
\end{iterbox}
\end{tcolorbox}
\caption{Prompt used during the speculation process in the GSM8k evaluation.}
\label{fig:synthesis prompt_gsm8k}
\end{figure}

\begin{figure}[H]
\begin{tcolorbox}[enhanced, colback=promptframe!3, colframe=promptframe, boxrule=0.7pt, arc=3pt, shadow={1pt}{-1pt}{0pt}{black!10}, left=3pt, right=3pt, top=3pt, bottom=3pt, title={\small\bfseries Validity Prompt $(p_{V})$ -- MAMV}, coltitle=white]
\scriptsize

\begin{iterbox}{Instructions}
You are following Hegel's dialectical method.
You have access to the Thesis and Antithesis.

\textbf{Your task:}
\begin{enumerate}[leftmargin=*, nosep]
    \item Read the Thesis and Antithesis below.
    \item Check whether Synthesis either:
    \begin{itemize}[nosep]
        \item Qualifies the contradictory statements so that Thesis and Antithesis unify without contradiction, OR
        \item Negates contradictory statements, explicitly listing which ones are negated and explaining why.
    \end{itemize}
    \item Check whether Synthesis presents a coherent, integrated perspective that resolves or clarifies these contradictions.
\end{enumerate}

Keeping all this in consideration, please assess whether the synthesis provided, derived from thesis and antithesis texts, adheres to the synthesis process. Reply with \texttt{yes} if it does and \texttt{no} if it does not.
\end{iterbox}

\begin{iterbox}{Input / Output}
\textbf{Source Text 1:} \{Thesis\}\\
\textbf{Source Text 2:} \{Antithesis\}\\
\textbf{Synthesized Text:} \{Synthesis\}\\[4pt]
\textcolor{exampleframe}{\textbf{Response (yes/no):}}
\end{iterbox}
\end{tcolorbox}
\caption{Validity assessment prompt in the MAMV paradigm.}
\label{fig:soundness}

\begin{tcolorbox}[enhanced, colback=promptframe!3, colframe=promptframe, boxrule=0.7pt, arc=3pt, shadow={1pt}{-1pt}{0pt}{black!10}, left=3pt, right=3pt, top=3pt, bottom=3pt, title={\small\bfseries Novelty Prompt $(p_N)$ -- MAMV}, coltitle=white]
\scriptsize

\begin{iterbox}{Instructions}
Please assess whether the source text is novel in comparison to the target text.
By novelty, we mean that the source contains new information that is not identical to the target.
Reply with \texttt{yes} if it is novel and \texttt{no} if it is not.
\end{iterbox}

\begin{iterbox}{Input / Output}
\textbf{Source Text:} \{Synthesis\}\\
\textbf{Target Text:} \{Thesis\}\\[4pt]
\textcolor{exampleframe}{\textbf{Response (yes/no):}}
\end{iterbox}
\end{tcolorbox}
\caption{Prompt utilized for the novelty assessment in the MAMV paradigm.}
\label{fig:novelty}
\end{figure}

\newpage
\section{Instances of Dialectical Iteration}
\label{app:iters}
In this section, we present the complete responses of the core self-reflective model (GPT-4o in our experiments), captured during three key moments: understanding, sublation, and speculation. These responses illustrate one full dialectical iteration during both scientific idea generation and mathematical reasoning tasks.

\begin{figure}[H]
\begin{tcolorbox}[enhanced, colback=promptframe!3, colframe=promptframe, boxrule=0.7pt, arc=3pt, shadow={1pt}{-1pt}{0pt}{black!10}, left=3pt, right=3pt, top=3pt, bottom=3pt, title={\small\bfseries Dialectical Iteration -- Idea Generation}, coltitle=white]
\scriptsize

\begin{iterbox}{Understanding ($\mathfrak{U}$)}
\textbf{Proposition $T_0$:}\; Energy is a continuous quantity.
\end{iterbox}

\begin{iterbox}{Sublation ($\mathfrak{A}$)}
\textcolor{phasered}{\textbf{Premises:}}
\begin{enumerate}[nosep]
    \item Quantum mechanics shows that energy is not always continuous, but can appear in quantized packets.
    \item Atomic and subatomic processes exhibit discrete energy levels.
\end{enumerate}
\textbf{Reasoning:}
These premises contradict the Thesis that energy is simply continuous, because the Thesis does not account for microscopic phenomena in which energy exchange occurs discretely.\\[2pt]
\textbf{Conclusion:} Energy is quantized in certain contexts, such as atomic and subatomic processes, where it exists in discrete packets rather than as a continuous entity.\\[2pt]
\textbf{Summary of Antithesis:} Energy cannot be understood as purely continuous; quantum phenomena require treating it as quantized in specific regimes.
\end{iterbox}

\begin{iterbox}{Speculation ($\mathfrak{S}$)}
\textcolor{exampleframe}{\textbf{Premises:}}
\begin{enumerate}[nosep]
    \item Energy can be described as continuous in classical physics, where it flows without discrete interruptions.
    \item In quantum mechanics, energy is quantized, existing in discrete packets such as photons or electron energy levels.
    \item Both descriptions depend on the scale and context of observation.
\end{enumerate}
\textbf{Reasoning:} The continuous description is a macroscopic approximation, while quantum mechanics reveals discreteness at microscopic scales. The two views are therefore not simply contradictory, but apply at different levels of description.\\[2pt]
\textbf{Conclusion:} Energy exhibits both continuous and quantized characteristics depending on context and scale.\\[2pt]
\textbf{Summary of Synthesis (Next Thesis):} Energy's duality as both continuous and quantized explains how classical and quantum mechanics complement each other across different scales.
\end{iterbox}
\end{tcolorbox}
\caption{An example of a dialectical iteration generated by GPT-4o for the idea generation experiment.}
\label{fig:example}
\end{figure}

\begin{figure}[H]
\begin{tcolorbox}[enhanced, colback=promptframe!3, colframe=promptframe, boxrule=0.7pt, arc=3pt, shadow={1pt}{-1pt}{0pt}{black!10}, left=3pt, right=3pt, top=3pt, bottom=3pt, title={\small\bfseries Dialectical Iteration -- Mathematical Reasoning (GSM8k)}, coltitle=white]
\scriptsize

\begin{iterbox}{Understanding ($\mathfrak{U}$) -- Proposition}
\textbf{Q:} John plans to sell all his toys and use the money to buy video games. He has $13$ lego sets and he sells them for $\$15$ each. He ends up buying $8$ video games for $\$20$ each and has $\$5$ left. How many lego sets does he still have?

\textbf{A:} He sold $13$ sets for $\$15$ each: $13 \times \$15 = \$195$. He bought $8$ games for $\$20$ each: $8 \times \$20 = \$160$. He has $\$5$ left, so total before buying: $\$160 + \$5 = \$165$. Since he made $\$195$ and spent $\$165$, he has no lego sets left. \textbf{The answer is 0 lego sets.}
\end{iterbox}

\begin{iterbox}{Sublation ($\mathfrak{A}$)}
\textcolor{phasered}{\textbf{Defects:}}
\begin{enumerate}[nosep]
    \item \emph{Misinterpretation:} The solution assumes John sold all his lego sets, but the problem says he plans to sell all his toys and asks how many lego sets remain.
    \item \emph{Logical error:} The solution calculates revenue from all $13$ sets instead of determining how many sets were actually sold from the amount spent and left over.
\end{enumerate}
Let $x$ be the number of lego sets sold. He spent $8 \times 20 = 160$ dollars and had $\$5$ left, so he earned $160 + 5 = 165$ dollars from selling lego sets.
Since each set sells for $\$15$, we have $15x = 165$, so $x = 11$.
He originally had $13$ sets, so $13 - 11 = 2$ remain. \textbf{The answer is 2 lego sets.}
\end{iterbox}

\begin{iterbox}{Speculation ($\mathfrak{S}$)}
The first solution correctly computes the total value of all $13$ lego sets, but incorrectly assumes that all of them were sold. The corrected solution resolves this by using the actual amount of money John needed for the games and the leftover amount. He needed $\$160$ for the games and had $\$5$ left, so he sold enough lego sets to make $\$165$. At $\$15$ per set, this means he sold $11$ sets. Since he started with $13$, he still has $2$ sets.\\[2pt]
\textbf{The answer is 2 lego sets.}
\end{iterbox}
\end{tcolorbox}
\caption{An example of a dialectical iteration generated by GPT-4o for a math problem from GSM8k.}
\label{fig:example_gsm}
\end{figure}

%% file: appendix_examples.tex
\section{Qualitative Results: Scientific Idea Generation}
\label{app:results}
This section provides examples from our qualitative experiment, i.e., the generation of new ideas across different domains. Each sequence traces one full dialectical trajectory: from an initial idea through successive sublation--speculation cycles, terminating when the MAMV criterion is met.


\begin{tcolorbox}[
  enhanced,
  colback=promptframe!3, colframe=promptframe,
  boxrule=0.7pt, arc=3pt,
  shadow={1pt}{-1pt}{0pt}{black!10},
  left=3pt, right=3pt, top=3pt, bottom=3pt,
  title={\small\bfseries Table 4: Topic 1 -- Mathematics (Annealing $\tau$, $\theta=0.5$)},
  coltitle=white,
]
\scriptsize

\begin{iterbox}{Iteration 0 -- Understanding ($\mathfrak{U}$)}
\textbf{Proposition $T_0$:}\; In mathematics, we can introduce multiple forms of infinity, some of which may be considered larger or smaller than others based on certain metrics, for example time, to allow for a sense of relativity when working with infinity.
\end{iterbox}

\begin{iterbox}{Iteration 1 \quad $\tau = 0.70$}
\textcolor{phasered}{\textbf{Sublation ($\mathfrak{A}$):}}
\textbf{Defects:} (1)~Does not fully define ``forms of infinity'' or how they can be differentiated. (2)~Does not account for the mathematical rigor required to compare infinities using metrics like time. (3)~Does not explain how relativity applies to infinity coherently.\\
\textbf{Opposition:} Infinity should be treated as a single abstract concept, compared through set theory and cardinality rather than physical metrics.

\tcblower
\textcolor{exampleframe}{\textbf{Speculation ($\mathfrak{S}$):}}
Infinity can be understood in multiple forms through set theory and cardinality, and the concept of relativity can be meaningfully applied within this framework to compare different sizes of infinity.
\end{iterbox}

\begin{iterbox}{Iteration 2 \quad $\tau = 0.42$}
\textcolor{phasered}{\textbf{Sublation ($\mathfrak{A}$):}}
\textbf{Defects:} (1)~Does not define how relativity can be rigorously applied within purely mathematical contexts. (2)~Fails to consider frameworks beyond set theory and cardinality. (3)~Use of time as a metric is unconventional and not sufficiently justified.\\
\textbf{Opposition:} Infinity should be understood through multiple mathematical frameworks; relativity must be clearly defined and justified.

\tcblower
\textcolor{exampleframe}{\textbf{Speculation ($\mathfrak{S}$):}}
Infinity can be understood through multiple frameworks (set theory, cardinality, topology, analysis), and the concept of relativity can be meaningfully applied to compare different sizes of infinity.
\end{iterbox}

\begin{iterbox}{Iteration 3 \quad $\tau = 0.25$}
\textcolor{phasered}{\textbf{Sublation ($\mathfrak{A}$):}}
\textbf{Defects:} (1)~Does not define how relativity applies rigorously to compare infinities. (2)~Fails to account for inherent differences between physical metrics and abstract mathematical concepts. (3)~Overlooks specific examples from topology and analysis.\\
\textbf{Opposition:} Precise definitions and rigorous application are essential; abstract concepts must be distinguished from physical metrics.

\tcblower
\textcolor{exampleframe}{\textbf{Speculation ($\mathfrak{S}$):}}
Infinity can be understood through multiple frameworks; precise definitions and rigorous application of principles are essential, and distinguishing abstract concepts from physical metrics is crucial.
\end{iterbox}

\begin{iterbox}{Iteration 4 \quad $\tau = 0.15$}
\textcolor{phasered}{\textbf{Sublation ($\mathfrak{A}$):}}
\textbf{Defects:} (1)~Does not fully define how relativity applies within mathematical frameworks. (2)~Does not account for potential incompatibility of physical metrics for measuring infinity. (3)~Lacks a clear explanation of how topology and analysis integrate.\\
\textbf{Opposition:} Physical metrics like time should be excluded from measuring infinity to maintain mathematical rigor.

\tcblower
\textcolor{exampleframe}{\textbf{Speculation ($\mathfrak{S}$):}}
Infinity can be understood through multiple frameworks; excluding physical metrics like time from the measurement of infinity is necessary to maintain mathematical rigor.
\end{iterbox}

\begin{iterbox}{Iteration 5 \quad $\tau = 0.09$}
\textcolor{phasered}{\textbf{Sublation ($\mathfrak{A}$):}}
\textbf{Defects:} (1)~Does not fully define how relativity can be applied. (2)~Excludes physical metrics without considering interdisciplinary insights. (3)~Lacks specific examples from topology and analysis. (4)~Does not address how to maintain rigor when integrating diverse frameworks.\\
\textbf{Opposition:} A comprehensive approach should include interdisciplinary insights, specific examples, and ensure rigor when integrating diverse frameworks.

\tcblower
\textcolor{exampleframe}{\textbf{Speculation ($\mathfrak{S}$) -- Final Idea:}}\\[2pt]
Infinity can be understood through multiple frameworks; interdisciplinary approaches, including insights from physical metrics, can enhance mathematical comprehension. Maintaining rigor when integrating diverse frameworks is crucial to avoid inconsistencies and ensure clarity.
\end{iterbox}

\vspace{2pt}
\centering{\small Process ended after 5 iterations \hfill {\color{exampleframe}Validity: \textbf{yes}} \quad {\color{promptframe}Novelty: \textbf{yes}}}
\label{tbl:math}
\end{tcolorbox}

\begin{tcolorbox}[
  enhanced,
  colback=promptframe!3, colframe=promptframe,
  boxrule=0.7pt, arc=3pt,
  shadow={1pt}{-1pt}{0pt}{black!10},
  left=3pt, right=3pt, top=3pt, bottom=3pt,
  title={\small\bfseries Table 5: Topic 2 -- Physics (Annealing $\tau$, $\theta=0.3$)},
  coltitle=white,
]
\scriptsize

\begin{iterbox}{Iteration 0 -- Understanding ($\mathfrak{U}$)}
\textbf{Proposition $T_0$:}\; Energy is a continuous entity.
\end{iterbox}

\begin{iterbox}{Iteration 1 \quad $\tau = 0.70$}
\textcolor{phasered}{\textbf{Sublation ($\mathfrak{A}$):}}
\textbf{Defects:} (1)~Does not account for the quantized nature of energy in quantum mechanics. (2)~Incompatible with discrete energy levels of electrons in atoms.\\
\textbf{Opposition:} Energy is quantized in certain contexts, existing in discrete packets rather than as a continuous entity.

\tcblower
\textcolor{exampleframe}{\textbf{Speculation ($\mathfrak{S}$):}}
Energy is a dual entity that can be continuous in classical contexts and quantized in quantum contexts, respecting both macroscopic and microscopic observations.
\end{iterbox}

\begin{iterbox}{Iteration 2 \quad $\tau = 0.51$}
\textcolor{phasered}{\textbf{Sublation ($\mathfrak{A}$):}}
\textbf{Defects:} (1)~Does not fully define the transition between continuous and quantized descriptions. (2)~Does not account for overlap phenomena. (3)~May oversimplify complex interactions.\\
\textbf{Opposition:} A more detailed examination of the transition between these descriptions is necessary for comprehensive understanding.

\tcblower
\textcolor{exampleframe}{\textbf{Speculation ($\mathfrak{S}$):}}
Energy is a dual entity, but its behavior at intermediate scales and in complex interactions requires nuanced understanding beyond simple categorization.
\end{iterbox}

\begin{iterbox}{Iteration 3 \quad $\tau = 0.38$}
\textcolor{phasered}{\textbf{Sublation ($\mathfrak{A}$):}}
\textbf{Defects:} (1)~``Intermediate scales'' is ambiguous. (2)~Does not account for quantum field theory advances. (3)~Does not provide specific failure examples.\\
\textbf{Opposition:} Energy should be understood through comprehensive frameworks like QFT, addressing transitions across all scales.

\tcblower
\textcolor{exampleframe}{\textbf{Speculation ($\mathfrak{S}$):}}
Energy is dual; modern theoretical frameworks like QFT can bridge descriptions across different scales. Accurate modeling requires clear definitions and specific examples.
\end{iterbox}

\begin{iterbox}{Iteration 4 \quad $\tau = 0.28$}
\textcolor{phasered}{\textbf{Sublation ($\mathfrak{A}$):}}
\textbf{Defects:} (1)~Lacks clear definition of intermediate scales. (2)~No specific examples of simplification failures. (3)~Assumes QFT can seamlessly bridge all scales without addressing limitations.\\
\textbf{Opposition:} Clear definitions, specific examples, and examination of QFT limitations at mesoscopic scales are required.

\tcblower
\textcolor{exampleframe}{\textbf{Speculation ($\mathfrak{S}$):}}
Energy is dual; QFT can bridge descriptions; accurate modeling requires clear definitions, specific examples, and addressing QFT's limitations at mesoscopic scales.
\end{iterbox}

\begin{iterbox}{Iteration 5 \quad $\tau = 0.21$}
\textcolor{phasered}{\textbf{Sublation ($\mathfrak{A}$):}}
\textbf{Defects:} (1)~Fails to provide practical definition of intermediate scales. (2)~Does not address how QFT applies to specific mesoscopic phenomena. (3)~Overly optimistic about QFT's bridging capability.\\
\textbf{Opposition:} Precise definitions and critical evaluation of QFT's limitations are essential for comprehensive understanding.

\tcblower
\textcolor{exampleframe}{\textbf{Speculation ($\mathfrak{S}$) -- Final Idea:}}\\[2pt]
Energy is dual (continuous classically, quantized quantum-mechanically). QFT can bridge these descriptions, but a critical evaluation of its limitations at mesoscopic scales is necessary for comprehensive understanding.
\end{iterbox}

\vspace{2pt}
\centering{\small Process ended after 5 iterations \hfill {\color{exampleframe}Validity: \textbf{yes}} \quad {\color{promptframe}Novelty: \textbf{yes}}}
\label{tbl:physics}
\end{tcolorbox}

\begin{tcolorbox}[
  enhanced,
  colback=promptframe!3, colframe=promptframe,
  boxrule=0.7pt, arc=3pt,
  shadow={1pt}{-1pt}{0pt}{black!10},
  left=3pt, right=3pt, top=3pt, bottom=3pt,
  title={\small\bfseries Table 6: Topic 3 -- Economics (Annealing $\tau$, $\theta=0.3$)},
  coltitle=white,
]
\scriptsize

\begin{iterbox}{Iteration 0 -- Understanding ($\mathfrak{U}$)}
\textbf{Proposition $T_0$:}\; According to Keynesian economics, during a recession the government should increase spending and cut taxes to stimulate demand. In times of economic boom, it should reduce spending and increase taxes to cool down the economy.
\end{iterbox}

\begin{iterbox}{Iteration 1 \quad $\tau = 0.70$}
\textcolor{phasered}{\textbf{Sublation ($\mathfrak{A}$):}}
\textbf{Defects:} (1)~Does not account for long-term impact on national debt and inflation. (2)~Assumes government intervention is always effective. (3)~Does not address potential inefficiencies in government spending.\\
\textbf{Opposition:} Government intervention should be carefully balanced and consider long-term impacts, limitations, and potential inefficiencies.

\tcblower
\textcolor{exampleframe}{\textbf{Speculation ($\mathfrak{S}$):}}
Government intervention can be valuable for stabilizing economic cycles, but it must be balanced with considerations for long-term impacts, efficiency, and external factors.
\end{iterbox}

\begin{iterbox}{Iteration 2 \quad $\tau = 0.51$}
\textcolor{phasered}{\textbf{Sublation ($\mathfrak{A}$):}}
\textbf{Defects:} (1)~Does not define ``efficient'' government spending. (2)~Does not account for crowding out private investment. (3)~Assumes effective balancing is always achievable despite political challenges.\\
\textbf{Opposition:} Government intervention should be minimized due to unclear efficiency definitions, crowding-out risks, and political challenges.

\tcblower
\textcolor{exampleframe}{\textbf{Speculation ($\mathfrak{S}$):}}
Government intervention must be balanced with considerations for efficiency, crowding-out effects, and political challenges to ensure sustainable stability.
\end{iterbox}

\begin{iterbox}{Iteration 3 \quad $\tau = 0.38$}
\textcolor{phasered}{\textbf{Sublation ($\mathfrak{A}$):}}
\textbf{Defects:} (1)~Still does not define spending efficiency. (2)~Does not consider positive impacts of public sector investment. (3)~Does not account for effective governance possibilities. (4)~Ignores external global factors.\\
\textbf{Opposition:} Government intervention can foster public sector innovation and social welfare; it requires clear definitions, successful governance recognition, and consideration of global factors.

\tcblower
\textcolor{exampleframe}{\textbf{Speculation ($\mathfrak{S}$):}}
Government intervention can stabilize cycles and foster innovation, balanced with efficiency, crowding-out effects, effective governance, and global factors.
\end{iterbox}

\begin{iterbox}{Iteration 4 \quad $\tau = 0.28$}
\textcolor{phasered}{\textbf{Sublation ($\mathfrak{A}$):}}
\textbf{Defects:} (1)~No actionable definition of efficiency. (2)~Claims about public sector innovation lack evidence. (3)~Underestimates complexities of political systems. (4)~Global factors are mentioned but not integrated.\\
\textbf{Opposition:} Government intervention should be critically assessed and limited due to these challenges.

\tcblower
\textcolor{exampleframe}{\textbf{Speculation ($\mathfrak{S}$) -- Final Idea:}}\\[2pt]
Government intervention can stabilize economic cycles and foster innovation, but must be balanced with long-term impacts, efficiency, crowding-out effects, effective governance, and global factors, supported by empirical evidence and a detailed framework.
\end{iterbox}

\vspace{2pt}
\centering{\small Process ended after 4 iterations \hfill {\color{exampleframe}Validity: \textbf{yes}} \quad {\color{phasered}Novelty: \textbf{no}}}
\label{tbl:economic}
\end{tcolorbox}

\begin{tcolorbox}[
  enhanced,
  colback=promptframe!3, colframe=promptframe,
  boxrule=0.7pt, arc=3pt,
  shadow={1pt}{-1pt}{0pt}{black!10},
  left=3pt, right=3pt, top=3pt, bottom=3pt,
  title={\small\bfseries Table 7: Topic 4 -- Philosophy (Annealing $\tau$, $\theta=0.3$)},
  coltitle=white,
]
\scriptsize

\begin{iterbox}{Iteration 0 -- Understanding ($\mathfrak{U}$)}
\textbf{Proposition $T_0$:}\; Existentialism emphasizes authenticity as a central value, positing that individuals must confront their freedom and take responsibility for crafting a meaningful life amidst an absurd and indifferent universe. Authenticity is achieved through self-awareness, rejection of external impositions, and commitment to one's values.
\end{iterbox}

\begin{iterbox}{Iteration 1 \quad $\tau = 0.70$}
\textcolor{phasered}{\textbf{Sublation ($\mathfrak{A}$):}}
\textbf{Defects:} (1)~Does not fully define ``authenticity'' or how it can be measured. (2)~Does not account for social and relational aspects of human existence. (3)~Assumes individuals can fully reject external impositions, which may not be feasible.\\
\textbf{Opposition:} Authenticity should be understood as a dynamic process influenced by social relationships and societal contexts.

\tcblower
\textcolor{exampleframe}{\textbf{Speculation ($\mathfrak{S}$):}}
Authenticity involves a dynamic interplay between personal values and societal influences, emphasizing ongoing negotiation rather than complete rejection of external impositions.
\end{iterbox}

\begin{iterbox}{Iteration 2 \quad $\tau = 0.51$}
\textcolor{phasered}{\textbf{Sublation ($\mathfrak{A}$):}}
\textbf{Defects:} (1)~Does not provide a clear, actionable definition. (2)~Does not address how external factors can dominate personal authenticity. (3)~Fails to provide a framework for balancing personal values and societal pressures.\\
\textbf{Opposition:} Authenticity should be clearly defined, and existentialism must offer a robust framework for navigating societal influence.

\tcblower
\textcolor{exampleframe}{\textbf{Speculation ($\mathfrak{S}$):}}
Authenticity is an evolving process shaped by internal self-awareness and external contexts, requiring balance supported by clear definitions and practical strategies.
\end{iterbox}

\begin{iterbox}{Iteration 3 \quad $\tau = 0.38$}
\textcolor{phasered}{\textbf{Sublation ($\mathfrak{A}$):}}
\textbf{Defects:} (1)~Clear definition may contradict the existentialist idea that authenticity is inherently subjective. (2)~Overlooks that societal influences can \emph{enhance} authenticity. (3)~Practical frameworks could become rigid and prescriptive.\\
\textbf{Opposition:} Authenticity is deeply personal and should not be rigidly defined; societal influences can support rather than undermine it; the pursuit should remain fluid.

\tcblower
\textcolor{exampleframe}{\textbf{Speculation ($\mathfrak{S}$):}}
Authenticity is a dynamic interplay between personal values and societal influences, emphasizing fluid, individualistic negotiation. It should not be rigidly defined; societal influences can both support and undermine one's sense of authenticity.
\end{iterbox}

\begin{iterbox}{Iteration 4 \quad $\tau = 0.28$}
\textcolor{phasered}{\textbf{Sublation ($\mathfrak{A}$):}}
\textbf{Defects:} (1)~Does not adequately define authenticity, leading to ambiguity. (2)~Lacks consideration of negative societal impacts (conformity). (3)~Does not address potential conflict between personal values and societal influences.\\
\textbf{Opposition:} Authenticity requires a clearer definition and thorough consideration of societal pressures, including conflicts with personal values.

\tcblower
\textcolor{exampleframe}{\textbf{Speculation ($\mathfrak{S}$):}}
Authenticity involves a dynamic interplay between personal values and societal influences, requiring both clarity in definition and flexibility in application, recognizing the unique journey of each individual.
\end{iterbox}

\begin{iterbox}{Iteration 5 \quad $\tau = 0.21$}
\textcolor{phasered}{\textbf{Sublation ($\mathfrak{A}$):}}
\textbf{Defects:} (1)~Does not account for societal norms becoming internalized and indistinguishable from personal values. (2)~Fails to provide criteria for achieving definitional clarity. (3)~Does not address subconscious influences. (4)~Assumes societal influences are external and separate.\\
\textbf{Opposition:} Authenticity is influenced by deeply internalized norms and subconscious factors, requiring a nuanced approach without assuming clear separation between internal and external influences.

\tcblower
\textcolor{exampleframe}{\textbf{Speculation ($\mathfrak{S}$) -- Final Idea:}}\\[2pt]
Authenticity involves a dynamic interplay between personal values, societal influences, and subconscious psychological factors, requiring both clarity in definition and flexibility in application. It remains fluid and individualistic, recognizing the complexities of maintaining authenticity in a social context.
\end{iterbox}

\vspace{2pt}
\centering{\small Process ended after 5 iterations \hfill {\color{exampleframe}Validity: \textbf{yes}} \quad {\color{promptframe}Novelty: \textbf{yes}}}
\label{tbl:philosophy}
\end{tcolorbox}